\documentclass[10pt,twocolumn,letterpaper]{article}
\usepackage[T1]{fontenc}
\usepackage{multirow}
\usepackage{capt-of} 
\usepackage[pagenumbers]{cvpr}

\usepackage{microtype}
\usepackage{booktabs}
\usepackage{threeparttable}
\usepackage{placeins}

\definecolor{cvprblue}{rgb}{0.21,0.49,0.74}
\usepackage[pagebackref,breaklinks,colorlinks,allcolors=cvprblue]{hyperref}

\def\confName{CVPR}
\def\confYear{2027}

\title{Any2Building: Unifying Point Clouds and Images for Building Reconstruction via Shared Spatial Queries}

\newcommand{\preprintauthors}{%
  \parbox[t]{\dimexpr\textwidth-2\tabcolsep\relax}{\centering
    Hongye~Hou\textsuperscript{1}, Zhuoxiao~Li\textsuperscript{1},
    Xiangyu~Liu\textsuperscript{1}, Tengxi~Wang\textsuperscript{1},
    Jing~Ou\textsuperscript{1}, Dongli~Wu\textsuperscript{1},
    Shuai~Zhang\textsuperscript{1}, Tongyan~Hua\textsuperscript{1},
    Wufan~Zhao\textsuperscript{1,\(\dagger\)}\par
    \vspace{0.35\baselineskip}
    \textsuperscript{1}The Hong Kong University of Science and Technology, Guangzhou, China.\par
    {\small\ttfamily hhou201@connect.hkust-gz.edu.cn\par}
    {\small\url{https://henryhoou.github.io/Any2Building/}\par}
  }%
}

\author{\preprintauthors}

\begin{document}
\ifdefined\pdfminorversion
  \pdfminorversion=7
\else
  
\fi

\twocolumn[{%
    \renewcommand\twocolumn[1][]{#1}
    \maketitle

    \begin{center}
        \includegraphics[width=0.93\textwidth]{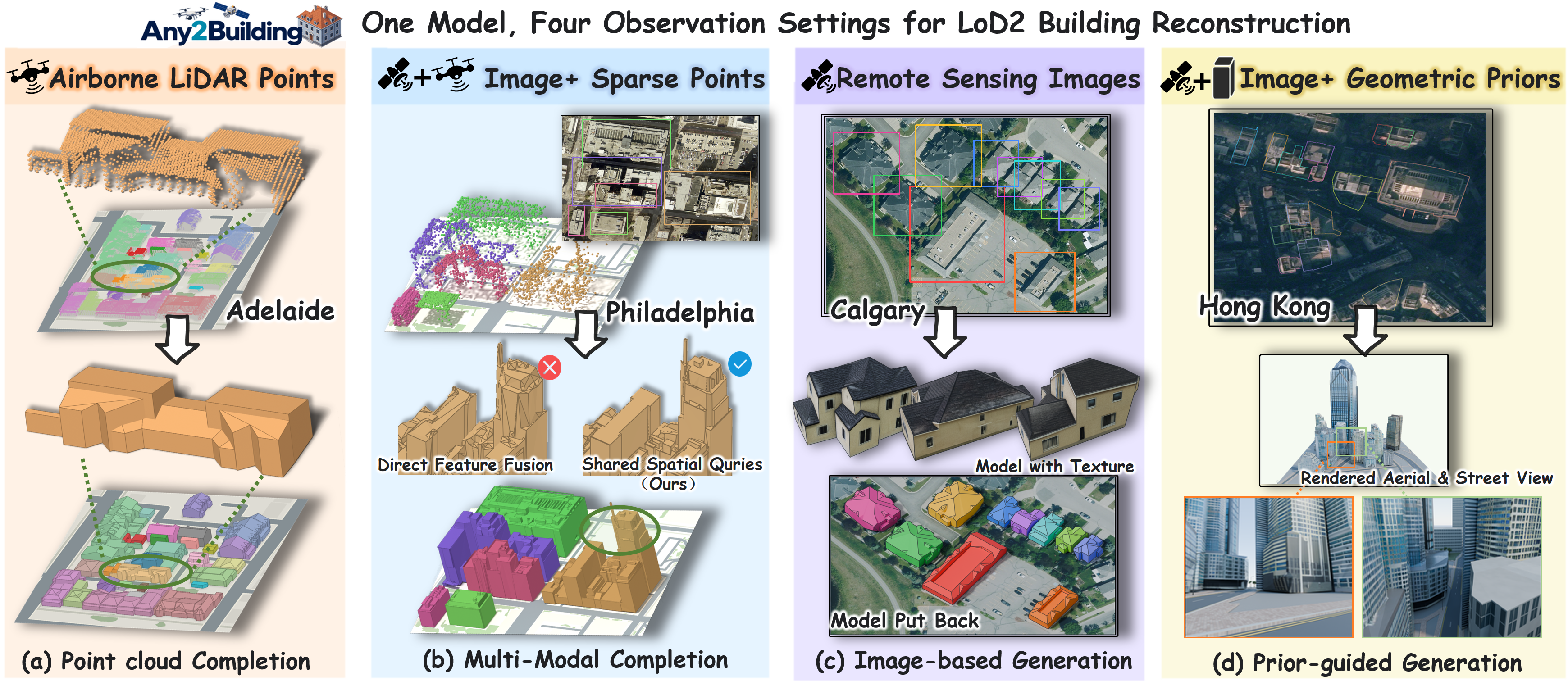}
        \captionof{figure}{%
            \textbf{Any2Building across heterogeneous observations.}
            A single jointly trained model accepts
            (a) partial point clouds,
            (b) sparse points with overhead imagery,
            (c) imagery cropped by instance bounding boxes, and
            (d) imagery with a coarse geometric prior.
            The reconstructed geometry supports compact LoD2 modeling and downstream texture synthesis.
        }
        \label{fig:teaser}
    \end{center}
    \vspace{0.5em}
}]

\begingroup
  \renewcommand{\thefootnote}{\fnsymbol{footnote}}
  \footnotetext[2]{Corresponding author.\par\vspace{0.4\baselineskip}}
\endgroup

\begin{abstract}
Abstracting structured buildings from limited observations
is essential for 3D urban modeling and digital twins.
Partial point clouds and overhead images of the same building provide
incomplete 3D surface measurements and visual cues to its footprint,
roof geometry, and rooftop structures, respectively.
This motivates a unified framework that aligns geometric and visual cues
in 3D and reconstructs buildings from either or both modalities.
Building on this connection, we present \textbf{Any2Building}, a unified model for Level of Detail 2 (LoD2) building reconstruction from partial point clouds, overhead images, or both, with optional coarse geometric priors. At its core is an occupancy-grounded shared spatial queries representation that organizes available observations within a common 3D grid. Supervised by complete building surface voxels, these queries connect observational evidence to target geometry, enabling a shared generator to learn structural priors across tasks and providing spatial guidance for recovering missing structure. A trained voxel conditional surface-generation and low-poly remeshing pipeline then converts the predicted structure into compact LoD2 building models. To support this framework, we introduce \textbf{BuildingWorld-Image}, comprising approximately one million paired samples of partial point clouds, overhead images, and LoD2 building models across 18 cities. 
Experiments on multiple benchmarks demonstrate that Any2Building outperforms existing methods in geometric reconstruction accuracy across unimodal and multimodal point-cloud completion, single-image building generation, and prior-assisted generation, while maintaining detailed and robust recovery when point cloud input modality is degraded.
\end{abstract}

\section{Introduction}
\label{sec:intro}

Structured 3D building models support urban digital twins~\cite{chen2024stylecity}
and applications such as urban understanding~\cite{zhang2026gs4city}
and visual localization~\cite{zhu2025lodlocv2}.
Level of Detail 2 (LoD2) modeling~\cite{huang2026buildingworld}
captures a building's spatial organization in a compact, structured
representation that preserves the geometry and spatial relationships
of its roof sections and detailed rooftop structures.
Building reconstruction often relies on limited observations, including
partial point clouds~\cite{hua2026buildanypoint,huang2025arcpro,liu2024point2building,huang2022city3d}
and overhead images~\cite{wei2023buildiff,jin2026satskylines}.
Point clouds provide partial geometric measurements, while
overhead images reveal appearance and layout but leave height
and hidden surfaces ambiguous.
Although these heterogeneous observations can originate
from the same building, point cloud based building
reconstruction and image-based building generation
are still largely treated as separate tasks.
This common origin motivates a unified reconstruction
framework that recovers LoD2 building
from either modality independently or their combination.

Recent building reconstruction methods have made progress
by drawing on powerful 3D generation
pipelines~\cite{ren2024xcube,xiang2025trellis,xiang2025trellis2,sam3d2025},
but their designs remain tailored to
individual observation settings and are difficult to
extend directly to complementary modalities.
BuildAnyPoint~\cite{hua2026buildanypoint}
reconstructs structured buildings from partial point clouds,
but its voxel-aligned conditioning
does not readily incorporate image evidence
as a spatially grounded 3D condition.
In prior-assisted image-based building
generation~\cite{jin2026satskylines},
extending noise-based prior injection to partial point clouds
can obscure fine structural cues.
Meanwhile, general-purpose 3D generators offer useful
inspiration through their efforts to unify heterogeneous
conditions within a shared
framework~\cite{cheng2023sdfusion,zhang2024clay,hunyuan2025omni,hu2026axolotl3d}.
Axolotl3D~\cite{hu2026axolotl3d}
and Hunyuan3D-Omni~\cite{hunyuan2025omni}
concatenate visual and geometric tokens for attention-based
conditioning, but rely on the generator to resolve
cross-modal structural relationships while inferring
missing geometry---a challenging joint inference problem
under incomplete observations.

This direct coupling is particularly challenging
in building reconstruction, where low-resolution
orthographic imagery provides limited height and facade
cues, yet reveals roof layouts and footprint boundaries
essential for LoD2 reconstruction.
However, direct feature fusion does not readily align
these visual cues with the fine-grained geometric
evidence in partial point clouds, as illustrated by the
direct-fusion example in Fig.~\ref{fig:teaser}(b).
The remaining challenge is to establish consistent
geometric correspondence between heterogeneous observations
and complete building structure,
enabling modality alignment and fusion
as well as task unification to promote positive transfer
across reconstruction tasks.

To address this challenge, we introduce \textbf{Any2Building},
a unified LoD2 building reconstruction model
that supports point cloud input and remote sensing
image inputs, together with height or box priors. 
Our main technical innovation is a set of shared queries initialized on a low-resolution 3D voxel grid. The available inputs are encoded as conditioning tokens, and shared
spatial queries aggregate their features at common 3D locations.
Supervision with complete building surface voxels encourages the queries
to capture the overall building structure, including unobserved regions,
before they condition the generator.
By relating different observations to the same underlying
geometry, the queries further guide the generator,
enabling shared structural learning across point-cloud
completion and image-based generation, together with optional coarse geometric priors. The generated structure latent is decoded into a voxel grid.
We find that these voxels alone suffice for continuous
mesh reconstruction by a trained surface generator,
without additional point or image conditioning.
Low-poly remeshing and texturing then produce compact Lod2 building assets.

To support learning across observation settings, we introduce
\textbf{BuildingWorld-Image}, an image extension of
BuildingWorld~\cite{huang2026buildingworld}.
It comprises approximately one million aligned triplets
of partial point clouds, overhead images, and LoD2 building
models across 18 cities.
These observations share the same geometric targets,
enabling structural learning and evaluation across input modalities.
Extensive experiments on multiple public datasets and BuildingWorld-Image dataset demonstrate that our method not only achieves superior performance on individual tasks but also benefits from complementary modalities when observations from point cloud modality are limited. 


Our contributions are summarized as follows.
\begin{itemize}
    \item We propose \textbf{Any2Building}, a unified model for LoD2
    building reconstruction from partial point clouds, overhead images,
    or both, with optional coarse geometric priors. The model combines
    complementary observations and promotes positive transfer across
    reconstruction tasks.

    \item We design a spatial conditioning module based on shared
    3D queries. Supervised by complete building voxels, these
    queries align heterogeneous observations with building geometry
    and provide spatial guidance for genrator.

    \item We introduce \textbf{BuildingWorld-Image}, a dataset containing
    approximately one million aligned triplets of partial point clouds,
    real overhead images, and LoD2 building models across 18 cities,
    supporting training and evaluation across observation settings.
\end{itemize}

\section{Related Work}
\label{sec:related}

\paragraph{Point cloud completion for building reconstruction.}
General point cloud completion methods recover missing geometry~\cite{hou2024dig} through
global feature encoding~\cite{pcn,foldingnet}, hierarchical
decoding~\cite{topnet,snowflakenet}, voxelized
representations~\cite{grnet}, and geometry-aware
transformers~\cite{pointr,adapointr,anchorformer}.
Multimodal methods, including EGIInet~\cite{xu2024egiinet},
DuInNet~\cite{liu2025duinnet}, and MAENet~\cite{liu2025maenet},
further incorporate paired images~\cite{hou2026benefit}.
These methods output point sets, requiring subsequent processing
for structured mesh reconstruction.

To recover coherent architectural structure,
PolyFit~\cite{nan2017polyfit} and City3D~\cite{huang2022city3d}
assemble planar primitives into compact meshes through constrained
face selection.
ArcPro~\cite{huang2025arcpro} predicts architectural programs encoding
building layers and their hierarchy, while
Point2Building~\cite{liu2024point2building} autoregressively generates
mesh vertices and faces from point clouds.
For existing meshes, RoLoPM~\cite{chen2023rolopm} combines offset
surface extraction, simplification, and geometric flow to produce
low-poly models while preserving visual fidelity.
BuildAnyPoint~\cite{hua2026buildanypoint} combines point-conditioned
diffusion with autoregressive mesh generation, while
UniCo~\cite{chen2026unico} uses learnable primitive proxies to
jointly predict point geometry and assembly-ready primitives
non-autoregressively. Both rely on observed 3D geometry through voxelized point
conditioning and shared point cloud features. Incorporating images therefore requires an additional
mechanism to spatially ground 2D appearance.

\paragraph{Image-based building generation.}
Recent image-to-3D models, including TRELLIS, TRELLIS.2,
and SAM 3D~\cite{xiang2025trellis,xiang2025trellis2,sam3d2025},
provide strong generative priors for recovering geometry
from visual observations.
At the building level, BuilDiff~\cite{wei2023buildiff}
generates point clouds through single-image-conditioned
diffusion, while SatSkylines~\cite{jin2026satskylines}
introduces geometric control by interpolating coarse-prior
latents with Gaussian noise.
However, extending this noise-space injection to partial
point clouds can obscure fine geometric cues, while coupling
geometric guidance with noise strength limits independent
control over local structure.
Related urban generation methods synthesize larger scenes
through compositional modeling~\cite{xie2024citydreamer,lin2023infinicity}
and satellite-conditioned
generation~\cite{hua2025sat2city,hua2026sat2city,qian2026sat3dgen,wang2025sat2realcity}.
Overhead appearance nevertheless leaves height and hidden
surfaces ambiguous, while coarse priors do not specify
detailed architectural geometry.

\paragraph{Geometrically conditioned 3D generation.}
Geometric conditioning enables more explicit control over
generated shapes.
SDFusion~\cite{cheng2023sdfusion} supports generation from
partial geometry and images, while CLAY~\cite{zhang2024clay}
accommodates diverse controls, including point clouds,
voxels, and bounding boxes.
SpaceControl~\cite{fedele2026spacecontrol} introduces spatial
constraints at test time without additional training.
Hunyuan3D-Omni~\cite{hunyuan2025omni} and
Axolotl3D~\cite{hu2026axolotl3d} concatenate visual and
geometric tokens for attention-based conditioning,
leaving the generator to infer how these heterogeneous
cues jointly constrain missing geometry.
This calls for a shared representation that
spatially organizes complementary observations and relates them to complete architectural
structure across input settings.
\section{Method}
\label{sec:method}

\begin{figure*}[t]
    \centering
    \includegraphics[width=0.95\textwidth]{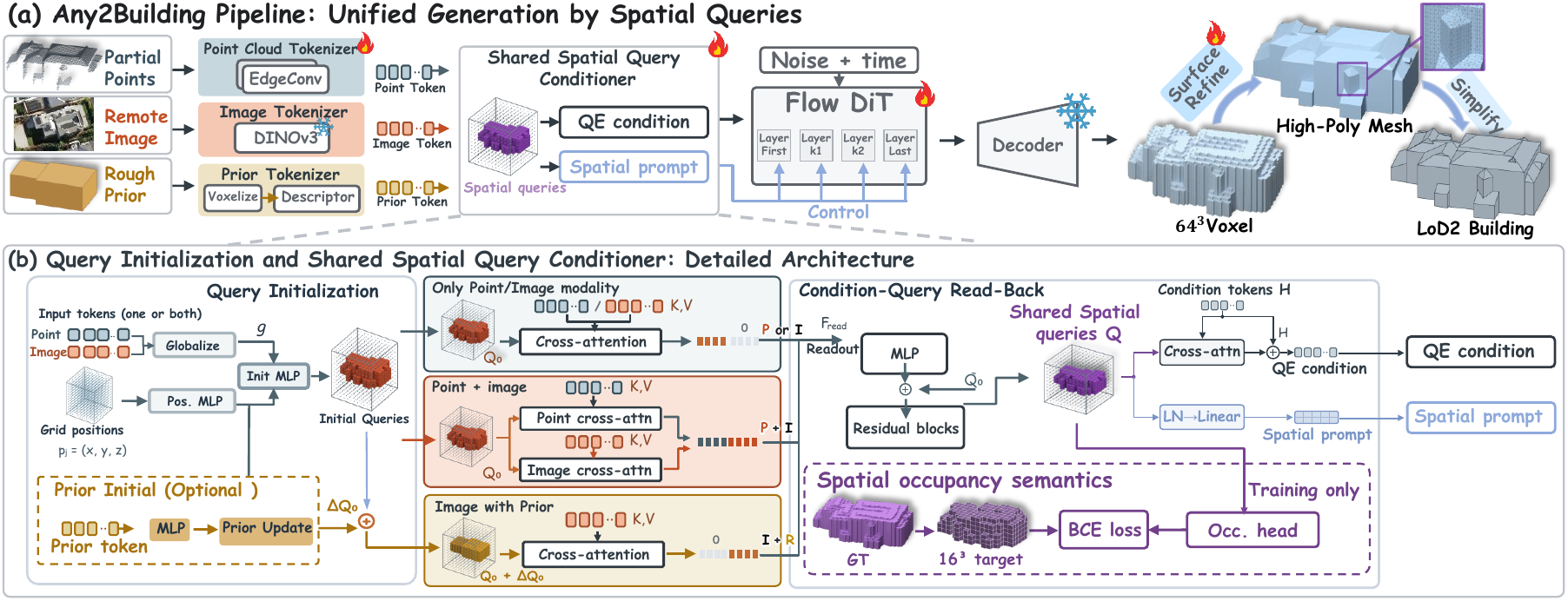}
    \caption{\textbf{Any2Building overview.}
    (a) Available observations and optional priors guide
    building generation through shared spatial queries,
    followed by surface generation and low-poly remeshing.
    (b) Queries on a coarse $16^3$ voxel grid aggregate
    observation features and optional priors.
    Voxel occupancy supervision encourages an intermediate
    representation of overall building structure, which
    conditions the DiT through QE tokens and spatial prompts.}
    \label{fig:pipeline}
\end{figure*}

Any2Building generates LoD2 buildings from a partial point
cloud $P$, an overhead image $I$, or both, with an optional
coarse geometric prior $R$.
Its central component is a set of shared spatial queries
that aggregate observations on a low-resolution 3D grid.
Supervised by complete building occupancy, these queries
form an intermediate representation of overall structure
before conditioning the generator.
As illustrated in Fig.~\ref{fig:pipeline}, the query features
guide a flow-matching DiT through query-enhanced (QE) tokens
and spatial prompts.
The generated structure is decoded into a $64^3$
surface-voxel grid and converted into a compact LoD2 mesh.
All geometric inputs, spatial queries, and output voxel
grids share a normalized building coordinate system
linked to the georeferenced overhead imagery.
This allows generated
buildings to be aligned with the source imagery and
mapped back to their real-world locations, even in
the image-only setting.

\subsection{Multimodal Input Tokenizers}
\label{sec:tokenizers}

We first encode the available observations for aggregation
by shared spatial queries.
Point clouds and images yield modality-specific tokens of
width $C=1024$, while optional geometric priors provide
descriptors for query initialization.
The point encoder uses four EdgeConv stages with
16-nearest-neighbor graphs in coordinate space.
Farthest-point sampling selects 2,048 and then 1,024 centers,
repeating samples when necessary.
The resulting center coordinates $P_c$ and local features
$H_c$ give
\begin{equation}
\begin{aligned}
(P_c,H_c) &= E_{\mathrm{EdgeConv}}(P),
T_P &= \phi_{\mathrm{feat}}(H_c)
      +\phi_{\mathrm{pos}}^P(P_c),
\end{aligned}
\label{eq:point_tokens}
\end{equation}
where the two MLPs encode local geometry and spatial
position, respectively, yielding
$T_P\in\mathbb{R}^{1024\times1024}$.
A frozen DINOv3 encoder~\cite{simeoni2025dinov3} extracts
image patch tokens $T_I\in\mathbb{R}^{N_I\times C}$.
The token sequences retain their modality-specific
organization; shared spatial queries associate them
with 3D locations.

LoD0 priors are 3D building bounding boxes; LoD1 priors
are building footprints extruded to their corresponding heights.
Each prior is voxelized into a binary $64^3$
surface-occupancy grid $V_R$, leaving interiors empty.
For each $4^3$ voxel block $\mathcal{B}_j$ corresponding
to a cell of the $16^3$ query grid, we compute
\begin{equation}
s_j=\max_{v\in\mathcal{B}_j}V_R(v),\;
b_j=\max_{v\in\mathcal{B}_j}B_R(v),\;
\rho_j=\frac{1}{64}\sum_{v\in\mathcal{B}_j}V_R(v),
\label{eq:prior_descriptors}
\end{equation}
where $B_R$ marks occupied voxels with at least one empty
six-connected neighbor; locations outside the grid are
treated as empty.
An MLP $\phi_R$ encodes the descriptor
$r_j=(s_j,b_j,\rho_j)$, representing surface occupancy
presence, boundary presence, and occupancy fraction.
For thin surface shells, $s_j$ and $b_j$ may coincide.

\subsection{Shared Spatial Query Conditioner}
\label{sec:query_conditioner}

We aggregate available observations into an intermediate
representation on a coarse 3D voxel grid, supervised by
complete-building surface occupancy.
The grid need represents both observed and missing regions,
providing structural guidance before the generator.

We initialize $N=16^3$ queries at fixed grid-cell centers
$\{\mathbf{p}_j\}_{j=1}^{N}$.
Mean-pooled, normalized modality features and availability
flags $\mathbf{a}=(a_P,a_I)$ form a global descriptor $g$,
with missing modalities contributing zero features.
The initial coarse queries are
\begin{equation}
q_j^0=
\phi_{\mathrm{init}}
\left(
[\phi_{\mathrm{pos}}^Q(\mathbf{p}_j),g]
\right),
\label{eq:query_init}
\end{equation}
where $[\cdot,\cdot]$ denotes concatenation.
When a prior is available, its descriptors update the
queries as
\begin{equation}
\widetilde q_j^0
=
q_j^0+\alpha\left[
s_j\operatorname{LN}
\left(\phi_{\mathrm{pos}}^Q(\mathbf{p}_j)\right)
+\phi_R(r_j)
\right],
\label{eq:prior_injection}
\end{equation}
where $\alpha$ scales the update.
Without a prior, $\widetilde Q^0=Q^0$.
All query locations are retained, allowing generation
beyond the occupied prior voxels.
Prior dropout during training reduces dependence on
this additional input.

The queries aggregate available modalities through cross-attention:
\begin{equation}
\begin{aligned}
A_m &=
\operatorname{Attn}_m
\left(
\operatorname{LN}(\widetilde Q^0),T_m,T_m
\right),
\quad m\in\{P,I\},\\
Q^r &= \widetilde Q^0+
\phi_f([A_P,A_I,\mathbf{a}]).
\end{aligned}
\label{eq:query_fusion}
\end{equation}
Missing modalities yield zero readouts, and $\mathbf{a}$ is
broadcast across queries.
Residual self-attention and feed-forward blocks, followed
by normalization, produce the spatial-query features $Q$. An auxiliary head predicts surface occupancy with the loss
\begin{equation}
\mathcal{L}_{\mathrm{occ}}
=
\frac{1}{N}\sum_{j=1}^{N}
\operatorname{BCE}
\left(
\sigma(w_{\mathrm{occ}}^\top q_j+b_{\mathrm{occ}}),y_j
\right).
\label{eq:query_occ}
\end{equation}
The targets $y_j$ are obtained by max-pooling the decoded
$64^3$ reference grid of the complete building to $16^3$.
They represent surface occupancy rather than filled interiors
and are shared across input settings.
This auxiliary loss only supervises $Q$ not the final voxels.

The query features $Q$ guide generation in two ways.
First, they enrich the base condition tokens $H$ through cross-attention.
For a single observation modality, $H=T_P$ or $H=T_I$.
For point--image inputs, cross-attention updates the point tokens with image
tokens and auxiliary completion features, preserving the point-token count
and width. The completion features come from 512 learnable queries distinct
from the $16^3$ spatial queries. The enriched condition is
\begin{equation}
H_{\mathrm{QE}}
=
H+\operatorname{Attn}_{\mathrm{back}}
\left(\operatorname{LN}(H),Q,Q\right).
\label{eq:qe_condition}
\end{equation}
Second, $Q$ provides spatial prompts that retain their 3D grid indices.

\begin{figure}[t]
    \centering
    \includegraphics[width=0.93\columnwidth]{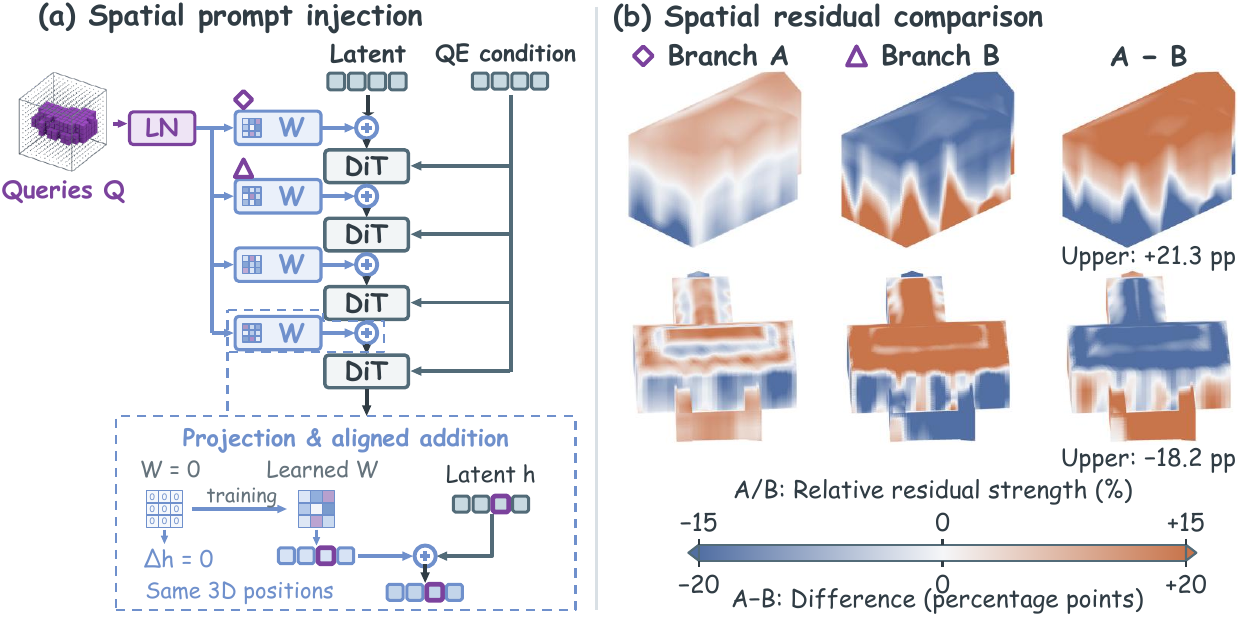}
\caption{\textbf{Spatial prompt injection.}
(a) Zero-initialized projections inject aligned residuals. QE tokens condition every block.
(b) A/B show RMS residual deviations from different spatial prompt block's surface mean. $A-B$ shows their difference.}
    \label{fig:spatial_prompt}
\end{figure}

\subsection{DiT with Spatial Prompts}
\label{sec:spatial_prompt}

The flow-matching DiT generates a $16^3$ structure latent,
conditioned on $H_{\mathrm{QE}}$ through cross-attention
in every block.
Selected blocks additionally receive aligned spatial prompts:
\begin{equation}
h_\ell
\leftarrow
h_\ell+\beta\,\mathcal{P}_\ell
\left(\operatorname{LN}(Q)\right),
\quad \ell\in\mathcal{S},
\label{eq:spatial_prompt}
\end{equation}
where $\mathcal{P}_\ell$ is a zero-initialized, layer-specific
linear projection and $\beta$ scales the residual.
Shared grid indexing aligns each query with its corresponding
latent location (Fig.~\ref{fig:spatial_prompt}). Figure 3(b) visualizes the spatial prompt residuals injected into two DiT blocks, labeled Branch A and Branch B. The layer-specific projections provide spatial guidance that emphasizes different parts of the building.

Given a complete-building latent $x_0$, Gaussian noise
$\epsilon$, and time $t\in[0,1]$, the flow path and target
velocity are
\begin{equation}
\begin{aligned}
x_t
&=(1-t)x_0+
[\sigma_{\min}+(1-\sigma_{\min})t]\epsilon,\\
u_t
&=(1-\sigma_{\min})\epsilon-x_0.
\end{aligned}
\label{eq:flow_path}
\end{equation}
Here $x_0$ denotes the clean target latent, and
$\sigma_{\min}$ specifies the residual noise level at $t=0$.
The training objective is
\begin{equation}
\begin{aligned}
\mathcal{L}
={}&
\mathbb{E}\left[
\left\|
v_\theta(x_t,t;H_{\mathrm{QE}},Q)-u_t
\right\|_2^2
\right]\\
&+\lambda_{\mathrm{occ}}\mathcal{L}_{\mathrm{occ}}
+\lambda_{\mathrm{geo}}\mathcal{L}_{\mathrm{geo}}.
\end{aligned}
\label{eq:training_loss}
\end{equation}
For voxel-level geometric supervision, clean latents are
estimated from $x_t$ and the predicted velocity using the
coefficients of Eq.~\ref{eq:flow_path}. The estimated latents are decoded into voxel grids, on which
$\mathcal{L}_{\mathrm{geo}}$ applies BCE and Dice supervision.

Training samples four input modes, $(P)$, $(I)$, $(P,I)$,
and $(I,R)$, which share the conditioner, generator, and
complete-building target representation.
QE tokens and spatial prompts are jointly dropped
during training for classifier-free guidance, which is a common training setting for DiT.

At inference, we integrate the flow from Gaussian noise at
$t=1$ to $t=0$ and denote the sampled endpoint by
$\widehat{x}_0$. The structure decoder maps this latent to
a $64^3$ voxel grid.

\subsection{LoD2 Construction}
\label{sec:lod2}

After obtaining the $64^3$ voxel grid $\widehat V$,
we generate local surface latents at its occupied
voxel coordinates $\Omega$.
We find that the predicted voxel geometry is sufficient for
surface generation without additional image or
input point-cloud features.
We therefore remove image conditioning from the
surface-stage structured-latent generator~\cite{xiang2025trellis2}
and train it on building data using
\begin{equation}
\mathcal{L}_{\mathrm{surf}}
=
\mathbb{E}_{z_0,\epsilon_z,t}
\left[
\left\|
v_{\psi}(z_t,t;\Omega)-u_t^{\mathrm{surf}}
\right\|_2^2
\right],
\label{eq:surface_loss}
\end{equation}
where $z_0$ denotes reference surface latents,
$z_t$ their interpolation with Gaussian noise $\epsilon_z$,
and $u_t^{\mathrm{surf}}=\partial z_t/\partial t$
the target flow velocity.
Training uses reference voxel coordinates, while
inference uses the predicted coordinates $\Omega$.

A pretrained shape decoder converts the generated
surface latents into a mesh, and
RoLoPM~\cite{chen2023rolopm} produces a compact
polygonal model retaining principal roof and
facade geometry.
This surface-construction pipeline is shared by all input modes.
For texturing, we synthesize plausible facade references using
ControlNet~\cite{zhang2023controlnet}, conditioned on untextured
mesh renderings and appearance descriptions inferred from the
overhead image and city name, then use pretrained
texture generator~\cite{xiang2025trellis2} to texture the low-poly mesh
with these references. The resulting LoD2 buildings can be mapped back into their
real-world scenes by reversing the normalization and applying
the associated georeferencing transforms.

\section{Experiments}

\subsection{Evaluation Protocol}
\begin{figure}[!b]
\centering
\includegraphics[width=0.95\columnwidth]{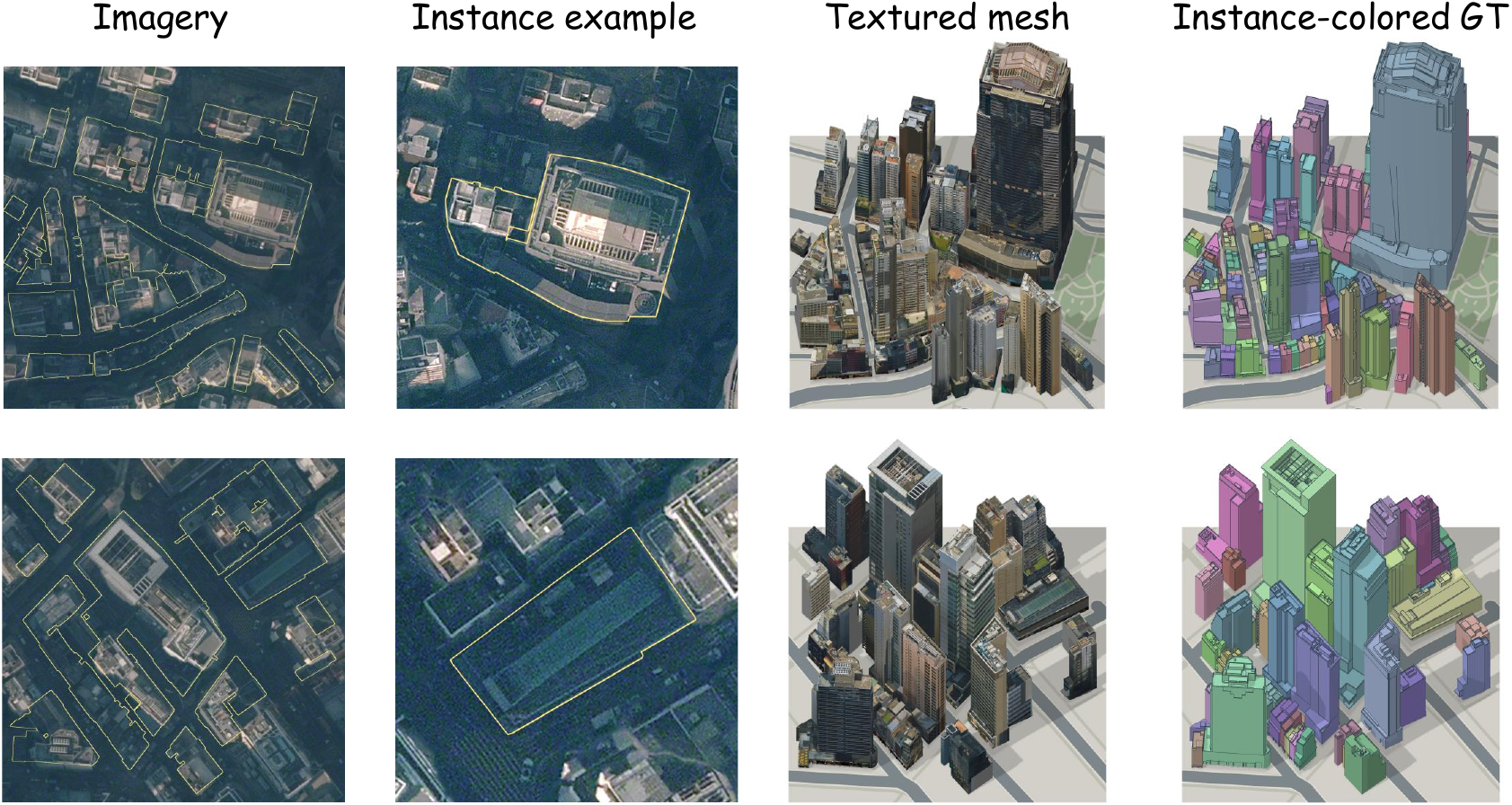}
\caption{\textbf{Real imagery with textured building ground truth in BuildingWorld-Image.} Remote-sensing observations, source textured meshes, and instance-colored geometry. }
\label{fig:textured-gt}
\end{figure}
\paragraph{Dataset.}
\begin{figure*}[t]
\centering
\includegraphics[width=\textwidth]{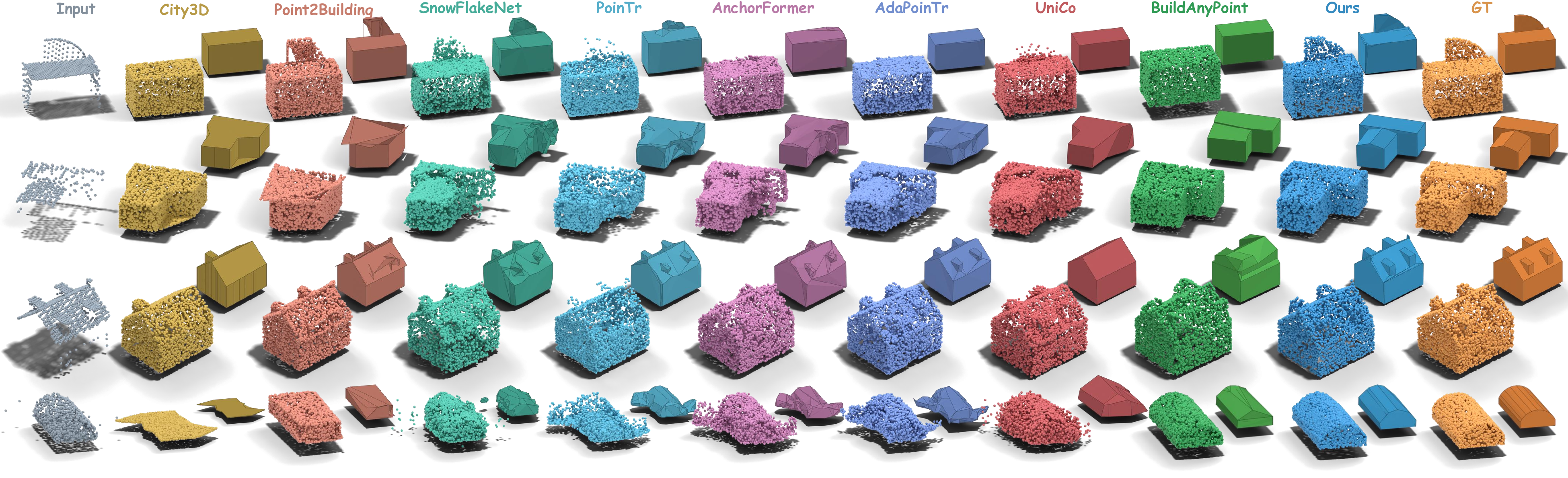}
\caption{\textbf{Completion and low-poly examples.}
Partial inputs, completed point sets, and compact geometric
outputs are shown alongside the reference. These selected
examples illustrate how point completion quality affects
the recovered architectural form.}
\label{fig:supp-completion}
\end{figure*}
Our BuildingWorld-Image contains 1,011,196 real overhead-image--low-poly mesh pairs across 18 cities before splitting, with approximately one million pairs used for training. The test set contains 2,000 buildings from unseen cities and held-out regions within training cities. Coarse priors are constructed from bounding boxes or height-extruded footprints. Subsets with textured ground-truth meshes support appearance evaluation through matched four-view renderings, with Hong Kong examples shown in Fig.~\ref{fig:textured-gt}. We also use Building-PCC~\cite{gao2024buildingpcc} and SatSkylines~\cite{jin2026satskylines} for point-cloud completion and prior-assisted generation comparisons, respectively. Dataset construction, splits, and evaluation protocols are detailed in Appendices A.

\paragraph{Evaluation settings and metrics.}
Within each input setting, methods share the evaluation frame and input preprocessing. Building-PCC follows the metric definitions and reporting scale of BuildAnyPoint~\cite{hua2026buildanypoint}. On BuildingWorld-Image, we report $\mathrm{CD}_{L_1}$, $\mathrm{CD}_{L_2}$, surface F-score at a normalized threshold of $0.02$, Earth Mover's Distance (EMD), uniformity (Uni.), and density-aware Chamfer distance (DCD). Both CD variants and EMD are scaled by $10^3$. The main experimental tables report F-scores as percentages. PSNR and SSIM evaluate appearance on the textured-reference subset. Voxel-support studies use voxel F1. Detailed metric definitions and alignment procedures are provided in the Appendices B.

\paragraph{Training and inference.}
Our model is trained for 200k iterations on eight NVIDIA A800 GPUs.
Inference uses the same GPU type, with 12 flow-sampling steps
and a guidance scale of 3.0.
Learned 3D generation baselines are fine-tuned on the corresponding
training split for 200k optimizer updates per independently
trained generative stage.

\subsection{Point-Cloud Completion}
\label{sec:point_completion}
\paragraph{Unimodal point-cloud completion.}
We first evaluate on Building-PCC~\cite{gao2024buildingpcc},
which includes LiDAR, SfM, and sparsely sampled point clouds.
We compare Any2Building with point completion baselines. Except for
City3D~\cite{huang2022city3d}, we train each method independently on this dataset.
Table~\ref{tab:completion-main} shows that Any2Building achieves
the best completion and low-poly accuracy, with PC and LP
F-scores of 96.12\% and 94.27\%, respectively. It also has the
smallest vertex and face count ratio errors, indicating close
agreement with reference mesh complexity.
Figure~\ref{fig:supp-completion} shows recovered roof structures
and coherent low-poly geometry from incomplete observations.

\begin{figure*}[!t]
\centering
\includegraphics[width=\textwidth]{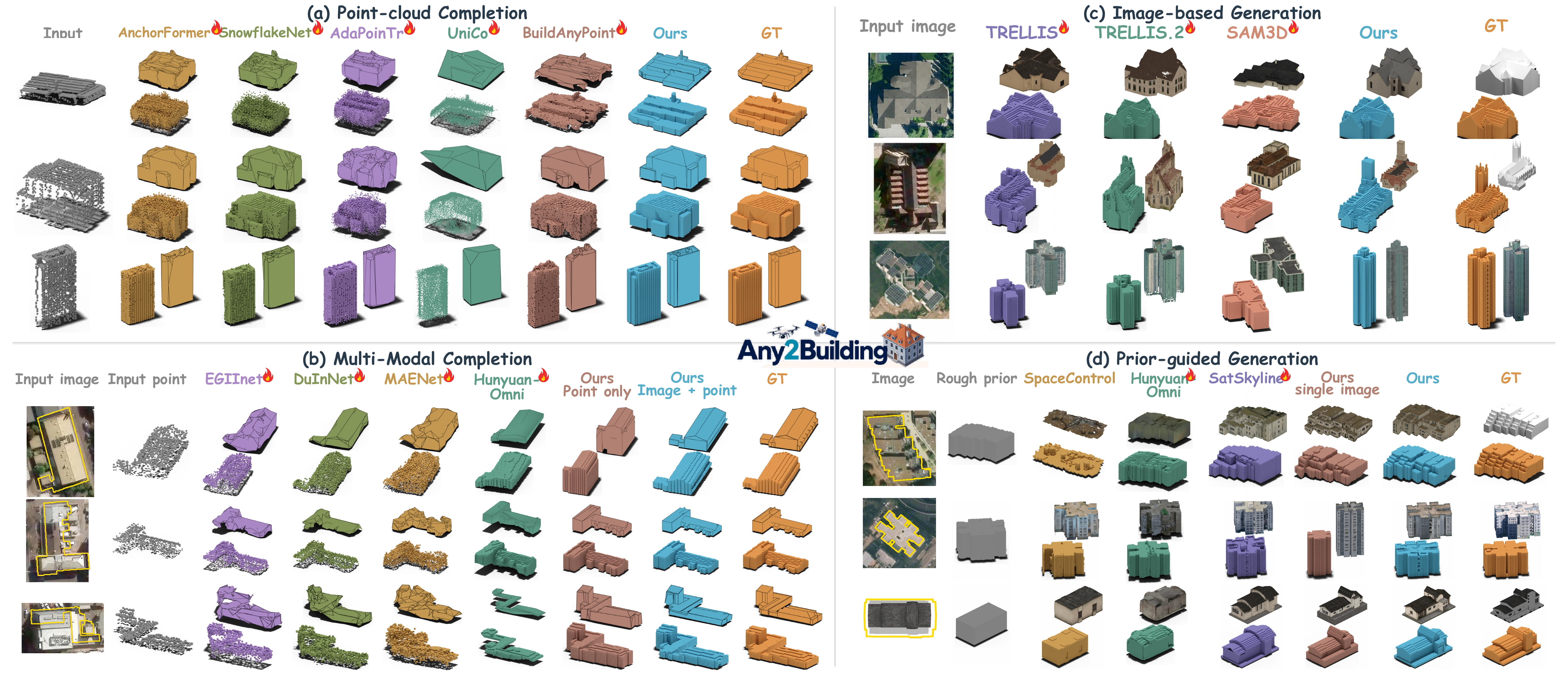}
\caption{\textbf{One model, four observation settings.} Any2Building reconstructs buildings from (a) partial points, (b) images with partial points, (c) a single overhead image, and (d) images with a rough prior. The examples highlight roof articulation, coherent building massing, and the use of complementary evidence. All compared models that support training (with fire logo) were trained under the corresponding task-specific experimental settings.}
\label{fig:pc_qualitative}
\end{figure*}

We further evaluate on BuildingWorld-Image.
As shown in Table~\ref{tab:completion-combined}(a), Any2Building
achieves the best results at both the point-cloud and low-poly
stages. Compared with AdaPoinTr, it reduces final low-poly
$\mathrm{CD}_{L_1}$ from 11.05 to 8.65 and increases the
F-score from 92.00\% to 97.19\%.
The qualitative comparisons in Fig.~\ref{fig:pc_qualitative}(a)
show more complete roof structures and cleaner building
boundaries, consistent with the improved geometric accuracy.

\begin{table}[t]
\centering
\caption{\textbf{Point-cloud completion and low-poly reconstruction on Building-PCC.} $r_V$ and $r_F$ are mean per-instance predicted-to-reference vertex and face count ratios. The parenthesized values $e_V$ and $e_F$ are their mean absolute deviations from 1.}
\label{tab:completion-main}
\footnotesize
\setlength{\tabcolsep}{2.5pt}
\resizebox{\columnwidth}{!}{%
\begin{tabular}{@{}lcccccccc@{}}
\toprule
& \multicolumn{4}{c}{Point cloud (PC)}
& \multicolumn{4}{c}{Low-poly mesh (LP)} \\
\cmidrule(lr){2-5}\cmidrule(lr){6-9}
Method & CD $\downarrow$ & F-score (\%) $\uparrow$
& EMD $\downarrow$ & Uni. $\downarrow$
& CD $\downarrow$ & F-score (\%) $\uparrow$
& $r_V\,(e_V{\downarrow})$ & $r_F\,(e_F{\downarrow})$ \\
\midrule
        City3D~\cite{huang2022city3d} & - & - & - & - & 1.49 & 45.43 & 1.62 (0.81) & 1.47 (0.63) \\
        Point2Building~\cite{liu2024point2building} & - & - & - & - & 1.20 & 52.92 & 1.29 (0.38) & 1.26 (0.41) \\
        FoldingNet~\cite{foldingnet} & 1.47 & 41.27 & 0.29 & 2.43 & 1.89 & 35.14 & 1.42 (0.61) & 1.58 (0.74) \\
        PCN~\cite{pcn} & 0.74 & 86.14 & 0.15 & 0.69 & 1.08 & 79.32 & 1.28 (0.39) & 1.36 (0.47) \\
        TopNet~\cite{topnet} & 0.97 & 25.43 & 0.35 & 0.61 & 1.60 & 19.28 & 1.35 (0.48) & 1.41 (0.52) \\
        GRNet~\cite{grnet} & 1.41 & 74.16 & 0.19 & 1.13 & 1.86 & 67.35 & 1.31 (0.44) & 1.33 (0.45) \\
        SnowflakeNet~\cite{snowflakenet} & 1.38 & 60.42 & 0.17 & 1.52 & 1.77 & 56.31 & 1.26 (0.36) & 1.29 (0.41) \\
        PoinTr~\cite{pointr} & 0.41 & 85.09 & 0.12 & 0.25 & \underline{0.55} & 77.46 & 1.18 (0.27) & 1.21 (0.32) \\
        AnchorFormer~\cite{anchorformer} & 0.39 & 82.35 & 0.13 & 1.27 & 0.60 & 75.18 & 1.22 (0.31) & 1.24 (0.35) \\
        AdaPoinTr~\cite{adapointr} & 0.44 & 70.24 & 0.15 & 1.49 & 0.64 & 65.37 & 1.15 (0.25) & 1.19 (0.29) \\
        UniCo~\cite{chen2026unico} & 0.43 & 73.16 & 0.14 & 1.44 & 0.63 & 68.29 & 0.76 (0.26) & 0.72 (0.30) \\
        BuildAnyPoint~\cite{hua2026buildanypoint} & \underline{0.35} & \underline{91.38} & \underline{0.10} & \underline{0.04} & 0.66 & \underline{80.14} & 0.83 (\underline{0.23}) & 0.80 (\underline{0.27}) \\
        \midrule
        Ours & \textbf{0.20} & \textbf{96.12} & \textbf{0.07} & \textbf{0.03} & \textbf{0.30} & \textbf{94.27} & 1.05 (\textbf{0.08}) & 1.03 (\textbf{0.07}) \\
\bottomrule
\end{tabular}%
}
\end{table}

\begin{table}[!t]
\centering
\caption{\textbf{Unimodal and multimodal point-cloud completion on BuildingWorld-Image.} P denotes partial points and I denotes an overhead image. Group (b) uses sparser partial observations than group (a). PC outputs precede surface generation and low-poly reconstruction. LP denotes final low-poly meshes.}
\label{tab:completion-combined}
\label{tab:supp-selected}
\label{tab:supp-multimodal}
\footnotesize
\setlength{\tabcolsep}{3pt}
\resizebox{\columnwidth}{!}{%
\begin{tabular}{@{}lccccccccc@{}}
\toprule
& & \multicolumn{2}{c}{Point cloud (PC)}
& \multicolumn{6}{c}{Low-poly mesh (LP)} \\
\cmidrule(lr){3-4}\cmidrule(lr){5-10}
Method & Input
& $\mathrm{CD}_{L_1}\downarrow$ & F-score (\%) $\uparrow$
& $\mathrm{CD}_{L_1}\downarrow$ & $\mathrm{CD}_{L_2}\downarrow$
& F-score (\%) $\uparrow$ & EMD $\downarrow$
& Uni. $\downarrow$ & DCD $\downarrow$ \\
\midrule
\multicolumn{10}{l}{\textit{(a) Unimodal point-cloud completion}} \\
PoinTr~\cite{pointr} & P & 11.24 & 91.00 & 12.68 & 0.79 & 88.00 & 92.42 & 0.34 & 0.52 \\
AnchorFormer~\cite{anchorformer} & P & 12.35 & 88.00 & 13.87 & 1.04 & 85.00 & 140.98 & 0.60 & 0.55 \\
SnowflakeNet~\cite{snowflakenet} & P & 10.42 & 93.00 & 11.69 & 0.78 & 91.00 & \underline{80.68} & \underline{0.08} & 0.47 \\
AdaPoinTr~\cite{adapointr} & P & \underline{9.81} & \underline{94.00} & \underline{11.05} & \underline{0.62} & \underline{92.00} & 90.57 & 0.14 & 0.48 \\
UniCo~\cite{chen2026unico} & P & 10.35 & 93.00 & 11.28 & 0.79 & 91.00 & 88.94 & 0.10 & 0.47 \\
BuildAnyPoint~\cite{hua2026buildanypoint} & P & 11.47 & 93.00 & 12.88 & 1.83 & \underline{92.00} & 86.56 & \underline{0.08} & \underline{0.46} \\
Ours & P & \textbf{9.25} & \textbf{96.25} & \textbf{8.65} & \textbf{0.40} & \textbf{97.19} & \textbf{68.12} & \textbf{0.04} & \textbf{0.30} \\
\midrule
\multicolumn{10}{l}{\textit{(b) Multimodal point-cloud completion}} \\
PoinTr~\cite{pointr} & P & 46.31 & 58.00 & 50.03 & 21.18 & 54.00 & 177.35 & 4.17 & 0.70 \\
AnchorFormer~\cite{anchorformer} & P & 45.72 & 61.00 & 49.28 & 24.25 & 58.00 & 206.22 & 0.57 & 0.69 \\
SnowflakeNet~\cite{snowflakenet} & P & 41.66 & 60.00 & 45.27 & 14.91 & 57.00 & 219.88 & 0.95 & 0.68 \\
AdaPoinTr~\cite{adapointr} & P & 54.28 & \underline{62.00} & 58.89 & 46.25 & \underline{59.00} & 207.00 & 0.75 & \underline{0.67} \\
UniCo~\cite{chen2026unico} & P & 53.61 & 50.00 & 58.42 & 43.51 & 55.00 & 172.48 & 0.12 & 0.71 \\
BuildAnyPoint~\cite{hua2026buildanypoint} & P & 52.94 & 40.00 & 57.20 & 20.64 & 58.00 & 166.13 & \underline{0.08} & 0.80 \\
\midrule
EGIInet~\cite{xu2024egiinet} & P+I & \underline{41.35} & 55.00 & 44.98 & 13.45 & 51.00 & 283.07 & 1.60 & 0.95 \\
DuInNet~\cite{liu2025duinnet} & P+I & 49.82 & 50.00 & 53.76 & 17.67 & 46.00 & 294.32 & 1.55 & 0.95 \\
Hunyuan3D-Omni~\cite{hunyuan2025omni} & P+I & 43.15 & 53.00 & \underline{44.52} & \underline{10.62} & 50.00 & 145.27 & 0.27 & 0.92 \\
MAENet~\cite{liu2025maenet} & P+I & 42.47 & 60.00 & 46.11 & 11.08 & 52.00 & \underline{134.51} & 0.20 & 0.94 \\
Ours & P+I & \textbf{27.53} & \textbf{75.45} & \textbf{26.00} & \textbf{10.07} & \textbf{77.41} & \textbf{119.74} & \textbf{0.06} & \textbf{0.66} \\
\bottomrule
\end{tabular}%
}
\end{table}

\paragraph{Multimodal point-cloud completion.}
Under sparser observations, all methods receive the same
partial points, with paired images additionally provided to
multimodal methods. Any2Building achieves the lowest low-poly
$\mathrm{CD}_{L_1}$ of 26.00 and the highest F-score of 77.41\%
in Table~\ref{tab:completion-combined}(b).
Figure~\ref{fig:pc_qualitative}(b) illustrates how image guidance helps recover missing building structures, including roof layouts and building components that are insufficiently constrained by sparse point observations.
The paired study in Table~\ref{tab:core-ablation}(b) isolates
the contribution of imagery under matched point observations.

\subsection{Image-Based Building Generation}
\label{sec:image_generation}

\begin{table}[!t]
\centering
\caption{\textbf{Image-based building generation on BuildingWorld-Image.} I denotes an overhead image and R denotes a coarse geometric prior. Backgrounds are retained in (a) and removed in (b). PSNR and SSIM compare matched four-view renderings of predicted and reference textured meshes.}
\label{tab:image-generation}
\footnotesize
\setlength{\tabcolsep}{2pt}
\resizebox{\columnwidth}{!}{%
\begin{tabular}{@{}lcrrrrrrrr@{}}
\toprule
& & \multicolumn{6}{c}{Geometry}
& \multicolumn{2}{c}{Appearance} \\
\cmidrule(lr){3-8}\cmidrule(lr){9-10}
Method & Input
& $\mathrm{CD}_{L_1}\downarrow$ & $\mathrm{CD}_{L_2}\downarrow$
& F-score (\%) $\uparrow$ & EMD $\downarrow$
& Uni. $\downarrow$ & DCD $\downarrow$
& PSNR (dB) $\uparrow$ & SSIM $\uparrow$ \\
\midrule
\multicolumn{10}{l}{\textit{(a) Single-image generation}} \\
TRELLIS~\cite{xiang2025trellis} & I & \underline{54.89} & 9.14 & \underline{49.20} & \underline{111.08} & \textbf{0.03} & \underline{0.66} & \underline{10.91} & 0.26 \\
TRELLIS.2~\cite{xiang2025trellis2} & I & 65.09 & 9.90 & 31.05 & 124.57 & \textbf{0.03} & 0.79 & 10.80 & \underline{0.28} \\
SAM 3D~\cite{sam3d2025} & I & 55.82 & \underline{8.53} & 44.34 & 117.59 & \underline{0.04} & 0.71 & 9.26 & 0.23 \\
Ours & I & \textbf{38.90} & \textbf{6.32} & \textbf{62.82} & \textbf{93.16} & \textbf{0.03} & \textbf{0.60} & \textbf{12.63} & \textbf{0.34} \\
\midrule
\multicolumn{10}{l}{\textit{(b) Prior-assisted generation}} \\
SpaceControl~\cite{fedele2026spacecontrol} & I+R & 34.77 & 3.83 & \underline{61.93} & 86.15 & \textbf{0.02} & 0.57 & \underline{11.65} & \underline{0.30} \\
Hunyuan3D-Omni~\cite{hunyuan2025omni} & I+R & \underline{32.46} & 3.51 & 60.82 & \underline{83.27} & 0.04 & \underline{0.55} & 11.38 & 0.29 \\
SatSkylines~\cite{jin2026satskylines} & I+R & 33.12 & \underline{3.42} & 59.62 & 84.69 & 0.05 & 0.56 & 11.21 & 0.28 \\
Ours & I+R & \textbf{17.23} & \textbf{1.53} & \textbf{78.64} & \textbf{57.82} & \underline{0.03} & \textbf{0.37} & \textbf{14.05} & \textbf{0.41} \\
\bottomrule
\end{tabular}%
}
\end{table}

\paragraph{Single remote-sensing image.}
We compare with TRELLIS~\cite{xiang2025trellis},
TRELLIS.2~\cite{xiang2025trellis2}, and SAM 3D~\cite{sam3d2025},
all fine-tuned on BuildingWorld-Image.
Any2Building reduces $\mathrm{CD}_{L_1}$ from 54.89 to 38.90
and increases the F-score from 49.20\% to 62.82\% compared with
TRELLIS (Table~\ref{tab:image-generation}(a)).
The full textured pipeline also achieves the highest PSNR
(12.63~dB) and SSIM (0.34).
Figure~\ref{fig:pc_qualitative}(c) illustrates the recovery of distinct building volumes and their spatial arrangement, including the separation of towers in the multi-volume example. We use bounding-box crops to retain building shadows and surrounding context and to avoid propagating errors from intermediate semantic segmentation.

\paragraph{Orthophoto with a geometric prior.}
Using the same orthophotos and coarse geometric priors, we compare Any2Building with SpaceControl~\cite{fedele2026spacecontrol}, Hunyuan3D-Omni~\cite{hunyuan2025omni}, and SatSkylines~\cite{jin2026satskylines}.
Any2Building achieves the lowest $\mathrm{CD}_{L_1}$ of 17.23 and the highest F-score of 78.64\% (Table~\ref{tab:image-generation}(b)).
The coarse priors constrain the overall building extent and massing but leave detailed roof geometry unspecified.
Figure~\ref{fig:pc_qualitative}(d) illustrates how our method recovers roof structures beyond those represented by the coarse priors.
Additional evaluations on 500 Skylines-50K assets show consistent advantages over the compared methods under both LoD0 and LoD1 priors (Appendix~C).

\subsection{Ablation Study}
\label{sec:ablation}

We evaluate spatial-query conditioning, shared learning, image complementarity, and surface generation on the full 2,000-building BuildingWorld-Image test set. Geometric metrics are computed on low-poly meshes.

\begin{table}[!t]
\centering
\renewcommand{\arraystretch}{1.0}
\caption{\textbf{Ablation of the unified generation pipeline.}
(a) Conditioning components and joint versus single-task training.
(b) Paired P and P+I inference under point degradation.
(c) Surface generation in the point-cloud-input setting. Both surface variants
use predicted voxel coordinates, and one additionally receives point-cloud
conditioning. F-scores are reported as percentages.}
\label{tab:core-ablation}
\footnotesize
\setlength{\tabcolsep}{2pt}

\begin{tabular*}{\linewidth}{@{\extracolsep{\fill}}lcccc@{}}
\toprule
\multicolumn{5}{l}{\textit{(a) Core components: LP F@2\% $\uparrow$}}\\
Variant & P & I & P+I & I+R\\
\midrule
\multicolumn{5}{l}{\textit{Spatial-query conditioning}}\\
Condition concat & 91.86 & 53.34 & 60.71 & 57.32\\
w/o occupancy supervision & 94.75 & 54.83 & 71.08 & 64.03\\
w/o query read-back & 93.78 & 51.17 & 72.44 & 70.38\\
w/o spatial prompts & 92.15 & 51.47 & 69.13 & 57.02\\
\midrule
Single-task training & 95.30 & 57.38 & 72.91 & 76.83\\
Full Any2Building & \textbf{97.19} & \textbf{62.82}
& \textbf{77.41} & \textbf{78.64}\\
\bottomrule
\end{tabular*}

\par\smallskip

\begin{tabular*}{\linewidth}{@{\extracolsep{\fill}}lrrrr@{}}
\toprule
\multicolumn{5}{l}{\textit{(b) Image contribution: LP F@2\% $\uparrow$}}\\
Degradation & Points & P (Sparse) & P (Sparse)+I & $\Delta F$ (pp)\\
\midrule
None & 2048 & 74.39 & \textbf{77.41} & +3.02\\
Random & 205 & 54.08 & \textbf{72.63} & +18.55\\
Random & 41 & 38.67 & \textbf{64.32} & +25.65\\
Spatial crop & 41 & 19.43 & \textbf{46.51} & +27.08\\
\bottomrule
\end{tabular*}

\par\smallskip

\resizebox{\linewidth}{!}{%
\begin{tabular}{@{}lcccccc@{}}
\toprule
\multicolumn{7}{l}{\textit{(c) Surface generation: final low-poly reconstruction}}\\
& \multicolumn{3}{c}{PolyFit~\cite{nan2017polyfit}}
& \multicolumn{3}{c}{RoLoPM~\cite{chen2023rolopm}}\\
\cmidrule(lr){2-4}\cmidrule(lr){5-7}
Variant
& Time (s)$\downarrow$ & $\mathrm{CD}_{L_1}\downarrow$ & F@2\%$\uparrow$
& Time (s)$\downarrow$ & $\mathrm{CD}_{L_1}\downarrow$ & F@2\%$\uparrow$\\
\midrule
Structure only
& \textbf{7.94} & 10.10 & 93.87
& \textbf{91.20} & 9.83 & 95.88\\
+ Surface (w/ point cond.)
& 16.12 & \underline{9.21} & \underline{94.58}
& 99.64 & \underline{8.72} & \underline{97.06}\\
+ Surface (w/o point cond.)
& 13.23 & \textbf{9.06} & \textbf{94.72}
& 97.57 & \textbf{8.65} & \textbf{97.19}\\
\bottomrule
\end{tabular}%
}

\end{table}

\begin{figure}[!t]
    \centering
    \includegraphics[width=0.95\columnwidth]{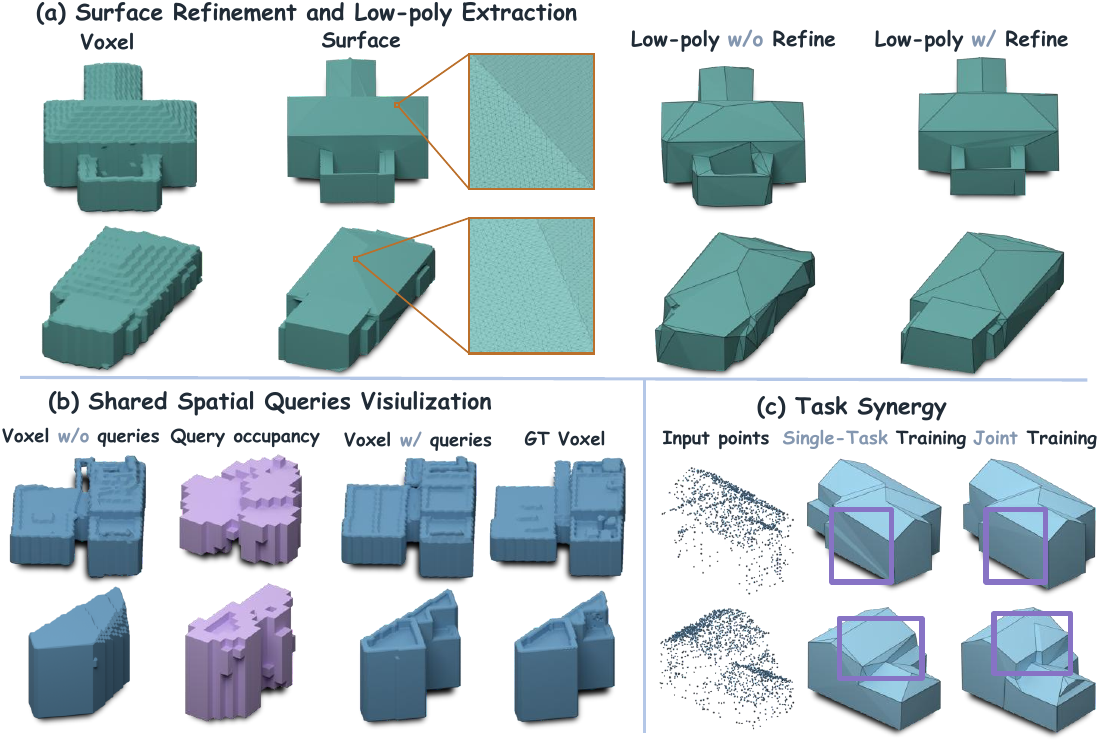}
    \caption{\textbf{From spatial queries to compact geometry.}
    (a) Surface generation reduces voxel staircase artifacts before low-poly remeshing.
    (b) Query and decoder occupancy alongside target surface voxels.
    (c) Predictions from single-task and jointly trained models.
    Quantitative component comparisons are reported in Table~\ref{tab:core-ablation}(a).}
    \label{fig:ablation}
\end{figure}

\paragraph{Spatial-query conditioning.}
Table~\ref{tab:core-ablation}(a) compares the full conditioning design with direct concatenation of the available input tokens, including prior tokens when present. We also remove occupancy supervision, query read-back, or spatial prompts individually and retrain each variant under the same training protocol. The full model achieves the highest F-score across all four settings. Without occupancy supervision, the image-only and prior-assisted F-scores decrease from 62.82\% to 54.83\% and from 78.64\% to 64.03\%, respectively, supporting the value of supervising the intermediate queries with complete building geometry. Removing either conditioning path also lowers performance. Without spatial prompts, the prior-assisted F-score drops to 57.02\%. Figure~\ref{fig:ablation}(b) visualizes query occupancy alongside the final decoder output.

\paragraph{Shared learning across tasks.}
We compare the joint model with one specialist per input setting, using the same number of training steps and the same inference inputs. Joint training samples the four input modes with equal probability and improves the final low-poly F-score by 1.89, 5.44, 4.50, and 1.81 percentage points for P, I, P+I, and I+R, respectively. The gains in P and I show that shared training also benefits reconstruction when only one observation modality is available at inference. Figure~\ref{fig:ablation}(c) compares predictions from single-task and jointly trained models. The jointly trained model predicts voxel structures that yield more regular low-poly meshes with higher structural fidelity, illustrating the benefits of shared learning across tasks.

\paragraph{Image contribution under sparse observations.}
We compare P and P+I inference using the same sparse points setting with imagery, With the original 2,048 point partial observations, adding imagery increases the F-score from 74.39\% to 77.41\%, a gain of 3.02 percentage points (Table~\ref{tab:core-ablation}(b)). With 41 retained points, adding imagery raises the F-score from 38.67\% to 64.32\% under random subsampling and from 19.43\% to 46.51\% after random spatial cropping followed by subsampling. Imagery thus provides greater benefits as geometric evidence becomes more sparse.
\paragraph{Surface generation and low-poly reconstruction.}
Surface refinement improves low-poly reconstruction with both
PolyFit and RoLoPM (Table~\ref{tab:core-ablation}(c)).
With RoLoPM, it increases the F-score from 95.88\% to 97.19\%
and reduces $\mathrm{CD}_{L_1}$ from 9.83 to 8.65.
Figure~\ref{fig:ablation}(a) illustrates the reduction in
voxel staircase artifacts.
Additional point-cloud conditioning slightly reduces accuracy
and increases runtime.
We therefore generate surfaces conditioned only on the
predicted voxel coordinates.


\section{Conclusion}
\label{sec:conclusion}

We presented Any2Building, a unified model for LoD2 building
reconstruction from partial point clouds, overhead images, or both.
The model also supports image-based reconstruction with coarse geometric
priors. Supervised by complete-building surface occupancy, our shared
spatial queries aggregate input features on a common 3D grid and
condition the same structure generator across input settings.
We generate surfaces from the predicted voxels alone and remesh them
into compact LoD2 models. We also introduced BuildingWorld-Image, comprising approximately one
million aligned triplets of partial point clouds, overhead images,
and LoD2 building meshes across 18 cities.
Experiments show improved reconstruction accuracy across unimodal and
multimodal point-cloud completion, single-image generation, and
prior-assisted generation. Ablations show that joint training improves
reconstruction from either modality alone, and that adding imagery
improves completion from sparse or spatially incomplete point clouds.
\FloatBarrier
\label{main-text-end}
\clearpage
{
    \small
    \bibliographystyle{ieeenat_fullname}
    \bibliography{main}
}

\clearpage
\appendix
\raggedbottom
\setcounter{figure}{0}
\setcounter{table}{0}
\renewcommand{\thefigure}{S\arabic{figure}}
\renewcommand{\thetable}{S\arabic{table}}
\ifdefined\theHfigure
\renewcommand{\theHfigure}{supp.\arabic{figure}}
\fi
\ifdefined\theHtable
\renewcommand{\theHtable}{supp.\arabic{table}}
\fi
\begin{center}
{\Large\bfseries Any2Building: Supplementary Material}
\end{center}

This supplementary material provides dataset details, implementation
and evaluation protocols, and additional analyses for Any2Building.
Appendix~\ref{sec:supp_dataset} describes dataset construction, quality
screening, splits, and image-based evaluation and rendering.
Appendix~\ref{sec:supp_protocol} presents implementation, geometric
metrics, ablation protocols, and surface-generation and low-poly
reconstruction results. Appendix~\ref{sec:supp_ablation} reports
encoder and query-resolution ablations and analyses of geometric
priors, alongside qualitative component comparisons and textured
visualizations.

\section{BuildingWorld-Image: Construction and Evaluation}
\label{sec:supp_dataset}
BuildingWorld-Image extends an existing geometric collection with geographically grounded visual observations. Its purpose is to connect building geometry, partial point measurements, and real overhead imagery within a common instance-level representation. This section details the data composition, construction and refinement procedures, partitioning, image-based evaluation protocols, and representative cases, including source textured meshes and instance-level correspondences (Fig.~\ref{fig:supp-textured-instances}).

\subsection{Composition and Geographic Coverage}
\label{sec:supp_composition}
Table~\ref{tab:dataset-cities} reports the tabulated BuildingWorld data pool
before train--test partitioning.
It contains 1,583,024 unique city-qualified geometric targets, including
1,011,196 targets paired with real overhead imagery after quality
filtering. This tabulated image subset spans 18 cities on four continents and
includes residential, industrial, and dense urban buildings. Perth
contributes 3,843 image--mesh pairs.

The image count is a subset of the geometry count, so the two columns
are not additive. Counts refer to source building targets rather than
augmented image views, point-cloud realizations, or prior variants.
Some source targets contain grouped structures. The paired subset
connects visual observations to geometry; the broader geometry pool
provides additional architectural examples.

\begin{table}[t]
\centering
\caption{\textbf{BuildingWorld data pool before partitioning.} Geometry counts unique
city-qualified low-poly targets. Image counts targets with a paired
overhead image after quality filtering, without tier quotas.
Image is a subset of Geometry; counts exclude augmentation multiplicity.
These are pre-partition counts, not final training-set sizes.}
\label{tab:dataset-cities}
\small
\setlength{\tabcolsep}{5pt}
\begin{tabular}{lrr}
\toprule
City & Geometry & Image \\
\midrule
Boston & 167,410 & 63,857 \\
Calgary & 322,649 & 224,190 \\
Cambridge & 17,041 & 11,191 \\
Cape Town & 274,598 & 125,202 \\
Edmonton & 370,673 & 246,540 \\
Greater Geelong & 886 & 596 \\
Hong Kong & 25,544 & 11,724 \\
Melbourne & 8,909 & 5,895 \\
Mississauga & 145,080 & 139,489 \\
Montreal & 24,866 & 24,652 \\
New York & 47,815 & 44,370 \\
Perth & 4,111 & 3,843 \\
Philadelphia & 2,907 & 2,117 \\
San Francisco & 91,359 & 65,583 \\
Tokyo & 34,507 & 10,998 \\
Toronto & 4,255 & 4,197 \\
Wellington & 15,213 & 13,165 \\
Yarra & 25,201 & 13,587 \\
\midrule
Total & 1,583,024 & 1,011,196 \\
\bottomrule
\end{tabular}
\end{table}

\subsection{Geographically Grounded Associations}
\paragraph{Recovering a common spatial reference.}
Source meshes use different coordinate systems, units, local origins, and axis conventions. We retain source metadata and recover the transform from local geometry to the geographic image frame before extracting instance observations. Coordinate projection, axis order, local offsets, and units are resolved in this transformation chain. The image affine mapping is updated after cropping, padding, and resizing so that pixel spacing continues to represent the correct physical extent. Meshes and partial points are placed in the same upright building coordinate frame for training. Source-to-geographic transforms remain attached to the association rather than being inferred from normalized geometry.

\paragraph{Retrieving overhead evidence.}
A building's geographic extent defines a context window in an orthophoto or remote-sensing mosaic. We retain the image-to-world transformation and available source metadata. Municipal orthophotos and mixed aerial/satellite basemaps are both used; consequently, we describe the collection as overhead remote-sensing imagery rather than assuming a uniform satellite sensor. The context preserves neighboring structures, while an instance mask focuses attention on the target.

\paragraph{Instance masks and paired conditions.}
We project the source building geometry into the geographic window to initialize an instance mask. RGB context, mask, and cutout are stored as related observations of the same target. Partial point clouds use available measurements or source-derived simulations according to their source pool. Coarse geometric conditions include bounding volumes and footprint-derived prisms. A prior that uses target-derived height is distinguished from one derived entirely from external observations; it is not treated as free information in image-only reconstruction. The projected instance mask is not supplied as an additional
condition in the single-image setting.

For quantitative evaluation, each target building is localized
using a 2D bounding box obtained by projecting its reference
geometry into the georeferenced image. The same bounding-box
crop is used for all compared methods. In the single-image
setting, background pixels inside the crop are retained.
The projected instance mask and its bounding box are distinct
annotations: storing a mask does not imply that background
removal is applied in every input setting.
Our image-based evaluation therefore assumes ground-truth-derived
instance localization and measures reconstruction given an
instance crop, rather than end-to-end building detection
and reconstruction. Although this is sufficient for single-building generation, we are also exploring, at this stage, the use of agents to automate the entire city-scale generation pipeline.

\paragraph{Geometric preparation.}
We remove invalid or empty geometric assets, standardize orientation, and encode the valid surfaces into sparse-structure and shape representations. Base-plane handling follows the source's conventions and the open-bottom training target; it is not a general watertightness repair. Geometry is normalized consistently with its associated point observations. 

\subsection{Quality Screening and Refinement}
\paragraph{Native image information.}
Quality screening checks native resolution, foreground coverage, exposure, texture, and orientation. Resolution tiers use the source image before resampling; enlarging a crop does not add evidence or improve its tier. Context and cutout variants share the same image window, padding, and display scale, with only the instance mask differing.

\paragraph{Coordinate, unit, and scale calibration.}
Systematic scale disagreement is first diagnosed in the coordinate conversion, rather than absorbed by a per-building image fit. Incorrect map projection, confusion between meters and feet, or a stale pixel-to-world scale after crop resizing can affect every target from a source. Our Boston repair restores the local offset and the source survey-foot convention. The inspected Cambridge export additionally requires a source-specific scale factor; it is not a universal correction for Cambridge data. These corrections belong to the shared source-to-world transform and are checked against geographic context before local alignment.

\paragraph{Different boundaries and height parallax.}
A visible roof boundary, an eave outline, and a ground footprint need not coincide. Projecting a source mesh into an image can also disagree with visible edges because an oblique acquisition displaces elevated surfaces relative to the ground, with different roof levels shifting differently. Occlusion and differences in acquisition date introduce further discrepancies. A single isotropic scale cannot generally resolve these effects. We therefore distinguish coordinate calibration from correspondence refinement, retain the meaning of each contour, and flag ambiguous roof--footprint matches instead of treating every residual as a scale error.

\paragraph{Shape-preserving boundary refinement.}
After geographic calibration, local refinement addresses residual contour displacement where the projected and visible boundaries are compatible. Our refinement proposal searches a bounded translation and isotropic scale around the registered contour. For contour point $p$ and its center $c$, the candidate transform is $T(p)=s(p-c)+c+t$. The objective combines orientation-compatible image-edge distance with penalties on translation $t$ and scale $s$, preserving the source polygon rather than reclassifying foreground pixels. The transformation preserves polygon components, vertex ordering, and corners, avoiding the erosion of roof structure that unconstrained image segmentation can introduce. We retain a proposal only when it improves the edge objective and remains inside the image; otherwise the registered prior is kept. Successive refinement passes record their cumulative scale, so a bound relative to an intermediate contour is not confused with a bound relative to the source geometry. Figure~\ref{fig:dataset-alignment} shows examples.

Refinement preserves vector-contour structure and retains the registered prior when a proposed update fails the acceptance criteria. No independent pixel annotations are available for this refinement, so the examples do not establish a mask-IoU improvement. Difficult cases involving occlusion, multiple roof levels, building groups, or an incorrect source contour remain subject to review.

\subsection{Splits and Representative Cases}
\label{sec:supp_splits}
The evaluation split contains 2,000 unique building targets in two groups.
The unseen-city group contains 250 buildings from each of Greater Geelong,
Philadelphia, Perth, and Toronto, for 1,000 targets. These four cities
are excluded from training across images, point clouds, geometry,
and derived observations.

The unseen-region group contains 100 buildings from each of Calgary,
Wellington, Tokyo, Montreal, Hong Kong, Cape Town, Edmonton,
Mississauga, New York, and Melbourne, for another 1,000 targets.
Contiguous geographic regions are reserved before selecting the test
buildings. All buildings in these regions are excluded from training,
not only the selected test targets. Spatial buffers cover at least the
image-crop extent to prevent overlap between training and test imagery.

The four held-out cities account for 10,753 image--mesh pairs in
Table~\ref{tab:dataset-cities}, leaving 1,000,443 pairs before region
and buffer exclusions. The resulting training set contains approximately
one million image--mesh pairs. Training uses eligible targets outside
the excluded cities, regions, and buffers. All image, point-cloud, and prior variants of a building
remain in the same partition. The training IDs, test IDs for both groups,
and geographic exclusion lists are frozen. The 18-city counts in
Table~\ref{tab:dataset-cities} precede these exclusions.
The query-resolution, surface-stage,
and paired degradation experiments use the full 2,000-building test set.

\begin{figure*}[p]
\centering
\includegraphics[width=\textwidth,height=.87\textheight,keepaspectratio]{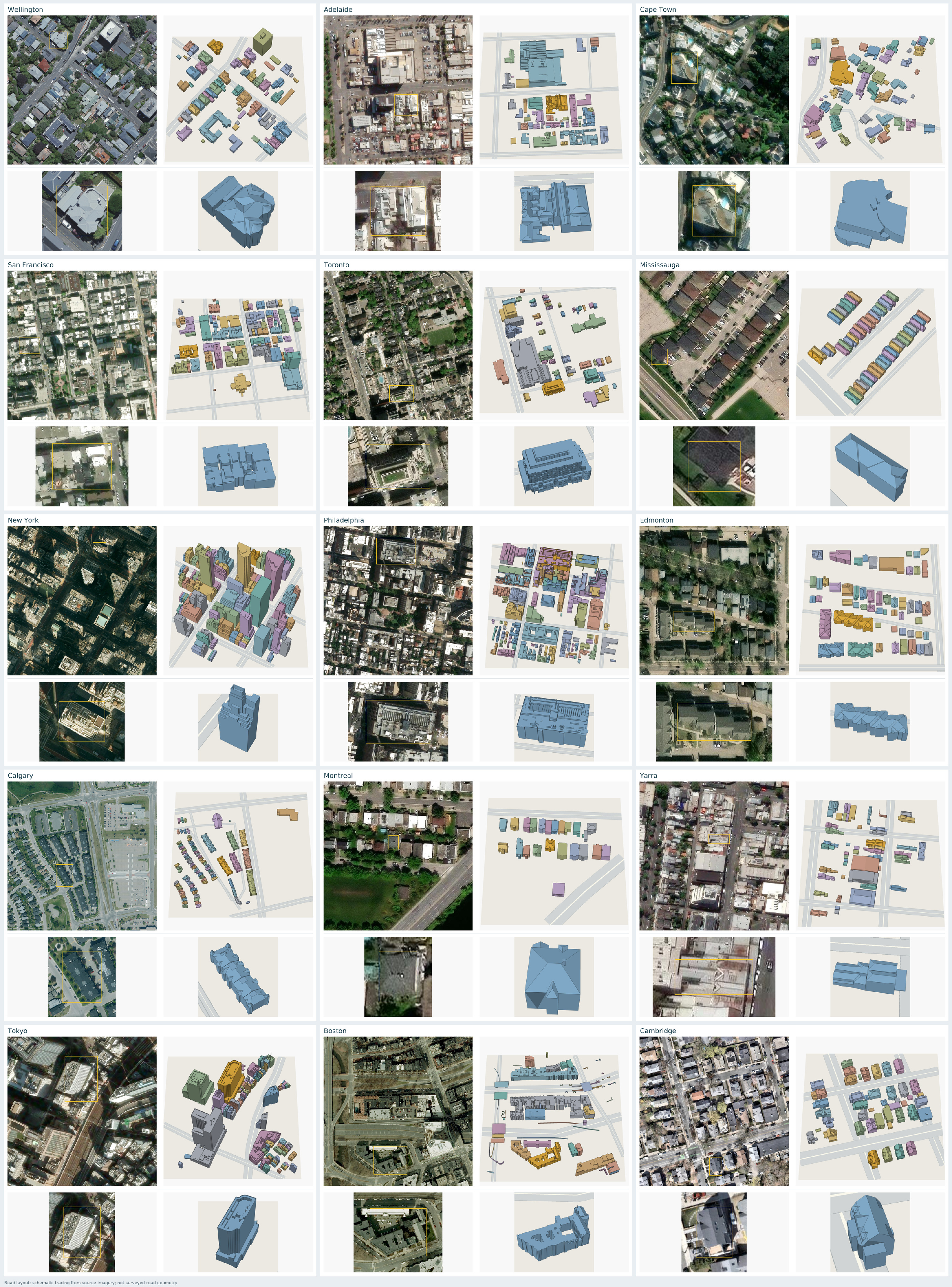}
\caption{\textbf{BuildingWorld-Image across 15 cities.} Each city panel pairs a geographic image window with source building geometry and shows a selected instance crop with its corresponding mesh. The examples illustrate geographic and architectural diversity; they are not restricted to the training subset. Road traces under rendered meshes provide schematic visual context and are not surveyed road annotations. Building geometry is unchanged.}
\label{fig:dataset-cases}
\end{figure*}

\begin{figure*}[p]
\centering
\includegraphics[width=\textwidth,height=.86\textheight,keepaspectratio]{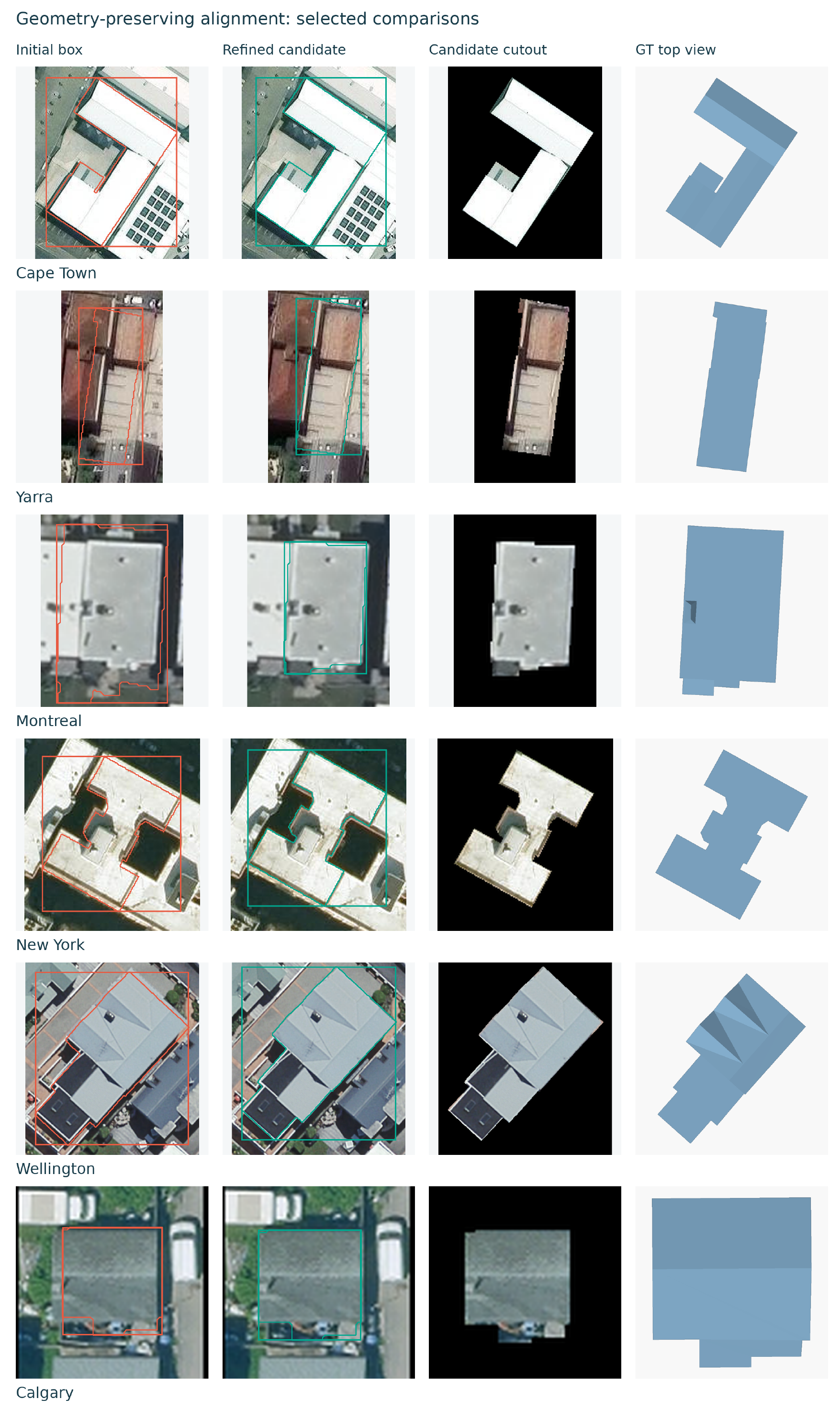}
\caption{\textbf{Preserving building structure during image alignment.} Six cities, with initial contours and boxes (red), registered candidates (green), candidate cutouts, and source-mesh top views. The shared image window preserves visual scale across the first three columns. Source meshes use a strict top-down view to check structure, not an estimated satellite camera. Local refinement preserves polygon structure; rows can retain an already registered input when no additional update is accepted. These curation examples are separate from the observations used to train the evaluated checkpoints; they do not provide independent mask annotations.}
\label{fig:dataset-alignment}
\end{figure*}

\begin{figure*}[p]
\centering
\includegraphics[width=\textwidth,height=.85\textheight,keepaspectratio]{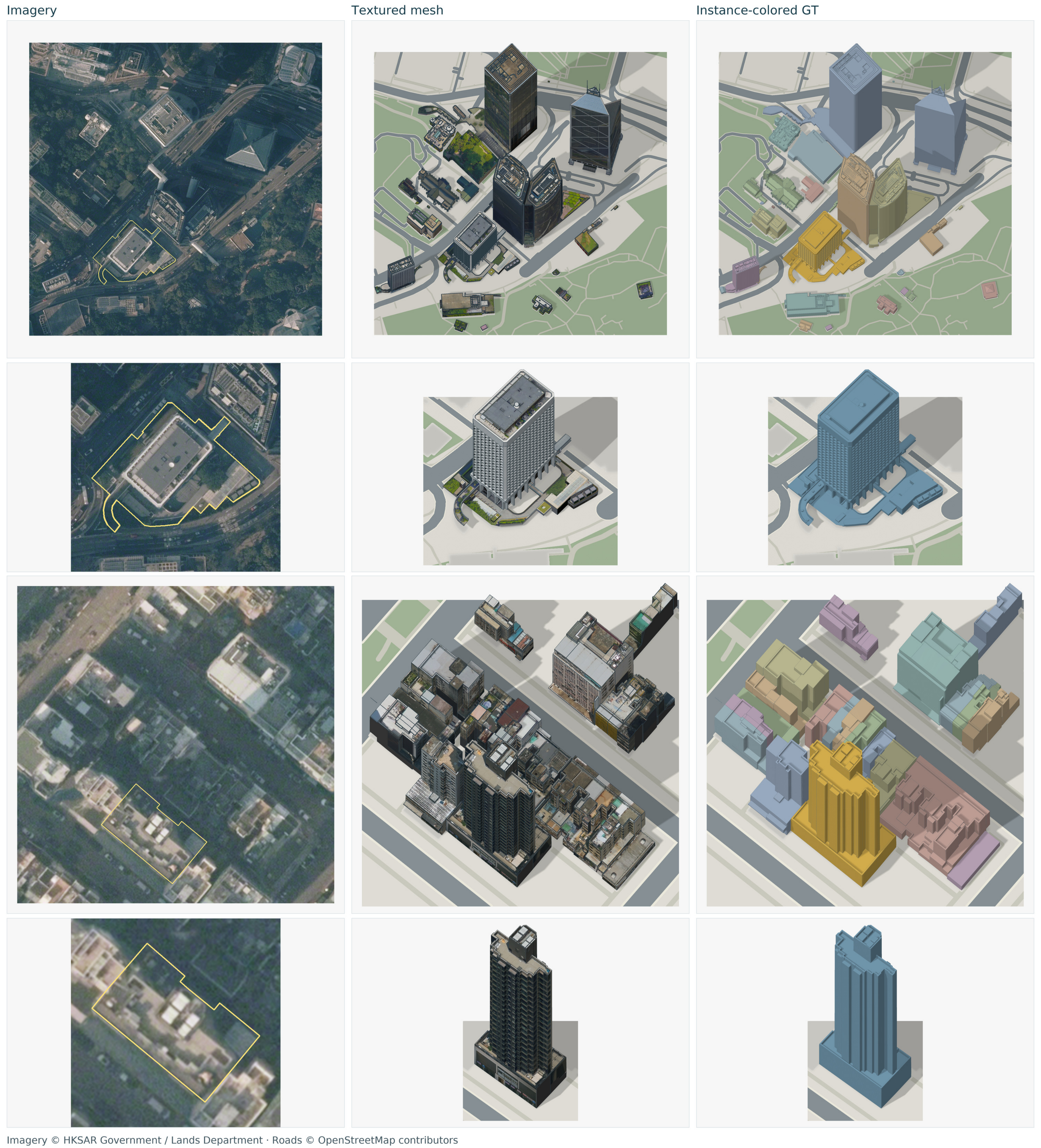}
\caption{\textbf{Textured ground truth from neighborhood to instance.} Each group shows the remote-sensing context and a selected building, with corresponding source textured mesh and instance-colored geometry. Geometry, camera, and framing are shared between the textured and colored mesh views. Source textures are retained; instance colors are display attributes, and road context is schematic. These are source references, not model predictions.}
\label{fig:supp-textured-instances}
\end{figure*}

Figure~\ref{fig:dataset-cases} illustrates associations across 15 cities,
which are not restricted to the training subset. It pairs geographic
context and source geometry with individual building observations.
Figures~\ref{fig:dataset-alignment} and
\ref{fig:supp-textured-instances} show alignment examples and textured
source meshes, respectively. These figures illustrate data construction
and source assets, rather than model predictions or a separate test set.

\subsection{Image-Based Evaluation and Rendering}
\label{sec:supp_image_eval}

The main-text image-generation table reports two input settings
with different image preprocessing protocols.
The single-image comparison uses real remote-sensing instance
crops with the background retained. It evaluates building-fine-tuned
TRELLIS, TRELLIS.2, SAM 3D Objects, and Any2Building using the same
input image and crop.
The prior-assisted setting instead uses background-removed instance
crops from orthophotos, reflecting OSM-assisted reconstruction
scenarios with available building footprints. Training in this
setting uses the corresponding background-removed images.
The comparison evaluates SpaceControl, Hunyuan3D-Omni, SatSkylines,
and Any2Building using the same image crop and coarse geometric
prior. Within each setting, all methods share the input
preprocessing, reference geometry, and evaluation frame.

\paragraph{Texture generation.}
All compared methods follow a reference-image-assisted
texturing workflow.
For each predicted untextured mesh, appearance-reference
images are synthesized and used to condition the corresponding
texture-generation module.
Each baseline retains its own texture generator.
The synthesized appearance references used for conditioning
are distinct from the ground-truth renderings used for evaluation.
Accordingly, appearance scores evaluate the complete
geometry-and-texturing pipelines, rather than isolating
the contribution of the geometry generator.

\paragraph{Appearance evaluation.}
PSNR and SSIM are evaluated on the subset with textured
reference geometry.
Following the four-view camera configuration of SatSkylines,
we render each predicted textured mesh and its reference
from yaw angles $\{0^\circ,90^\circ,180^\circ,270^\circ\}$,
with a fixed pitch of $30^\circ$, field of view of $40^\circ$,
and camera radius of $2$ in the normalized rendering frame.
Predictions and references use identical camera poses,
rendering resolution, background, and lighting.

We compute PSNR and SSIM between corresponding predicted
and reference renderings, average the four view scores
for each building, and then average across buildings.
RGB values are normalized to $[0,1]$.
These metrics compare rendered textured meshes rather than
renderings against the input remote-sensing image, and
therefore reflect both geometric and appearance differences.

\clearpage
\clearpage
\clearpage

\clearpage

\section{Implementation, Evaluation, and Reconstruction}
\label{sec:supp_protocol}

\subsection{Architecture, Training, and Inference}
\label{sec:supp_training}
The point encoder uses four EdgeConv stages with 16-neighbor
aggregation and farthest-point reductions to 2,048 and 1,024 centers.
It returns 1,024 spatially indexed tokens of width 1,024. The frozen
DINOv3 encoder processes images at $512\times512$. The spatial-query
lattice contains $16^3$ locations. Query features are projected into
a 30-block diffusion transformer of hidden width 1,536. Spatial prompts
enter blocks 0, 10, 20, and 29 (zero-based indices), while query-enhanced
condition tokens supply cross-attention throughout the transformer.

The shared training configuration samples point-only (P), image-only
(I), point--image (P+I), and image-plus-prior (I+R) modes with equal
probability. Point degradation precedes resampling to the encoder input
size; repeated input rows do not constitute additional observations.
The query occupancy loss has weight 0.1. Voxel-level geometric supervision uses
the frozen sparse-structure decoder with a BCE/Dice auxiliary loss
weighted by 0.02. Query-enhanced tokens and spatial prompts are dropped
together with probability 0.1 for classifier-free guidance.

At inference, the surface generator operates at the predicted occupied voxel
coordinates, without additional image or point-cloud features. It is
therefore conditioned on the predicted structure, even though direct
observation conditioning is removed.

\paragraph{Baseline adaptation.}
We initialize the learned 3D generation baselines from their
released pretrained checkpoints and fine-tune them on the
training split corresponding to each experiment.
Each independently trained generative stage receives 200,000
optimizer updates, using the task-specific input preprocessing
described in Appendix~\ref{sec:supp_image_eval}.
For Hunyuan3D-Omni, we fine-tune the shape diffusion model
and unified control encoder jointly as a single shape-generation
stage. The update budget refers to optimizer steps rather than
inference sampling steps.

\paragraph{Baseline availability.}
BuilDiff and Axolotl3D are discussed as related work but
are excluded from the quantitative comparisons because
we could not obtain runnable official implementations
as of November 15, 2026.
BuilDiff's official repository contained only a README
and license, while the Axolotl3D project page did not
provide a public implementation or checkpoint download.

\subsection{Geometric Metrics and Output Representations}
\label{sec:supp_metrics}
The Building-PCC comparison and encoder ablation follow
the metric definitions and reporting scale of
BuildAnyPoint~\cite{hua2026buildanypoint}.
Its supplementary material defines CD as the sum of the two
directional mean nearest-neighbor L1 (Manhattan) distances,
using 16,384 points per cloud.
The geometric distance definitions and scaling detailed below
apply to BuildingWorld-Image.

In the completion comparisons, PC denotes completed point
sets before surface generation and low-poly reconstruction,
whereas LP denotes final low-poly meshes.
Mesh-based geometric metrics are computed from surface-sampled
points.

Reference meshes are sampled uniformly by surface area with 16,384
points. A common evaluation frame uses the reference centroid and
maximum radius for every method. For predicted points $X$ and reference
points $Y$, define $d(x,Y)=\min_{y\in Y}\|x-y\|_2$. We compute
\begin{equation}
\mathrm{CD}_{L1}=\frac12\left[
\frac{1}{|X|}\sum_{x\in X}d(x,Y)+
\frac{1}{|Y|}\sum_{y\in Y}d(y,X)\right],
\end{equation}
and
\begin{equation}
\mathrm{CD}_{L2}=
\frac{1}{|X|}\sum_{x\in X}d(x,Y)^2+
\frac{1}{|Y|}\sum_{y\in Y}d(y,X)^2.
\end{equation}
Precision $P_\tau$ is the fraction of predicted points within $\tau$ of
the reference, and recall $R_\tau$ is defined in the reverse direction.
The surface F-score is
\begin{equation}
F_\tau=\frac{2P_\tau R_\tau}{P_\tau+R_\tau},
\qquad \tau\in\{0.01,0.02\},
\end{equation}
with $F_\tau=0$ when $P_\tau+R_\tau=0$. Unless stated otherwise,
reported surface F-scores use $\tau=0.02$.

EMD uses exact bijective matching on 1,024 sampled points and averages
two sampling seeds. CD and EMD in this evaluation are multiplied by
$10^3$. For point outputs, CD and F-score use the native prediction
point count. Uniformity and DCD use 16,384 points, with replacement
only when necessary. Uniformity is the normalized variance of neighbor
counts within radius 0.05 at 100 query points, averaged over three
seeds. DCD uses $\alpha=1000$ and count exponent 1.

Voxel F1 in Appendix~\ref{sec:supp_query_design} measures overlap
between predicted and reference surface-support voxels. These voxels
represent building surfaces, not filled interiors. Voxel F1 and surface
F-score measure different quantities and are not numerically interchangeable.

\paragraph{Low-poly complexity.}
For the Building-PCC low-poly results in
Table~\ref{tab:completion-main}, we compare predicted and
reference vertex and face counts.
For building $i$ and count type $c\in\{V,F\}$, let
$n^{\mathrm{pred}}_{c,i}$ and $n^{\mathrm{ref}}_{c,i}$
denote the corresponding counts, and define
$\rho_{c,i}=n^{\mathrm{pred}}_{c,i}/n^{\mathrm{ref}}_{c,i}$.
Over $N$ evaluated buildings, we report
{
\begin{equation}
\begin{aligned}
r_c &= \frac{1}{N}\sum_{i=1}^{N}\rho_{c,i},\\
e_c &= \frac{1}{N}\sum_{i=1}^{N}
       \left|\rho_{c,i}-1\right|,
\qquad c\in\{V,F\}.
\end{aligned}
\end{equation}
}
Each table entry reports $r_c$ with $e_c$ in parentheses.
The error is averaged over individual buildings rather than
computed from the aggregate ratio, preventing overestimation
and underestimation across buildings from cancelling out.

\subsection{Ablation Protocols}
\label{sec:supp_ablation_protocols}

\paragraph{Evaluation scope.}
All three studies in Table~\ref{tab:core-ablation} use the full 2,000-building BuildingWorld-Image test set. Geometric metrics are evaluated on the final low-poly meshes produced by the reconstruction procedure specified for each study, using the shared evaluation frame and metric definitions in Appendix~\ref{sec:supp_metrics}. F-score uses a normalized distance threshold of 0.02, and $\mathrm{CD}_{L1}$ is reported with a scale factor of $10^3$. Component variants are compared within each input setting in panel (a); the paired contribution of imagery is evaluated separately in panel (b). In the surface-generation study, both the structure-only and surface-stage variants are converted to low-poly meshes using the indicated backend before evaluation.

\paragraph{Spatial-query variants.}
The \emph{condition concat} baseline replaces the spatial-query conditioning design with direct concatenation of the available input tokens. Point and image tokens are included according to the input setting, and encoded prior tokens are also concatenated in the I+R setting. The resulting sequence directly conditions the DiT, without query read-back or spatial-prompt injection. This baseline therefore compares the complete spatial-query conditioning design with direct token conditioning while retaining access to the available input modalities and priors.

The remaining variants remove one component at a time. The \emph{w/o occupancy supervision} variant omits the auxiliary query occupancy loss while retaining the query representation and the remaining generation objectives. The \emph{w/o query read-back} variant uses the base condition tokens $H$ directly, i.e., $H_{\mathrm{QE}}=H$, while retaining query occupancy supervision and spatial prompts. The \emph{w/o spatial prompts} variant removes the aligned residual injection into the DiT while retaining query-enhanced condition tokens and occupancy supervision. Each variant is retrained using the same training protocol; the components are not simply disabled in the trained full model at inference.

\paragraph{Joint and single-task training.}
We train one specialist for each of P, I, P+I, and I+R, using the corresponding architecture and inference inputs. Each specialist and the joint model receive the same number of optimizer updates. The joint model samples the four input modes with equal probability, whereas each specialist uses only its designated mode. The comparison therefore matches the total update count per model rather than the number of training exposures to each mode. At evaluation, the joint model and the corresponding specialist receive the same type of input; joint training does not introduce extra inference modalities into the P-only or I-only comparisons.

\paragraph{Paired point degradation.}
The image-contribution study uses the same trained checkpoint for P and P+I inference. The \emph{None} setting uses the original 2,048-point partial observation without additional degradation; it does not denote a complete building point cloud. Random degradation retains 205 or 41 points from this observation. Spatial degradation adapts the ShapeNet-55 cropping scheme used by AdaPoinTr~\cite{adapointr}. A randomly selected spatial region is first removed, after which the remaining observations are subsampled to 41 points. Thus, spatial removal precedes subsampling, and the 41-point input is not obtained by directly cropping away all but 41 points. Comparing this setting with random subsampling to the same count examines the role of spatial coverage under equally sparse observations.

For every comparison, P and P+I use identical retained points, normalization, and generation noise; P+I additionally receives the paired overhead image. Degradation is applied before resampling to the encoder input size, so repeated samples do not add geometric observations. The reported point counts refer to the retained observations rather than the length of the resampled encoder input.

Surface-stage variants and their point-conditioning comparison are described in Appendix~\ref{sec:supp_lowpoly}; conversion and timing settings are specified in Appendix~\ref{sec:supp_conversion}.

\subsection{Low-Poly Conversion and Timing Protocol}
\label{sec:supp_conversion}
Point-output baselines use a fixed $64^3$ marching-cubes adapter before
low-poly conversion. The structure-only variant uses its predicted
support mesh, and the complete pipeline uses the structure-conditioned
surface generator. BuildAnyPoint supplies its native completion-to-mesh
result, which is repeated as a reference in both backend groups of
Table~\ref{tab:supp-lowpoly}.

RoLoPM uses a screen-size setting of 100 and an intermediate budget of
128k triangles. PolyFit uses 20k surface samples with fixed plane-detection
and optimization settings, including one common coarse-plane retry upon
failure. For the PolyFit adapter, open-bottom surface-stage outputs receive
4k bottom-support samples derived from their predicted minimum height.
These representation-specific adapters remain fixed within the comparison.

Reported time sums cached GPU inference and measured CPU conversion
using eight cores. It excludes initialization, I/O, cleanup/export,
metric computation, and evaluation alignment. It is an aggregate of
recorded components rather than a newly measured end-to-end wall-clock
time. Results are reported in Appendix~\ref{sec:supp_lowpoly}.

\subsection{Surface Generation and Low-Poly Reconstruction}
\label{sec:supp_lowpoly}
Table~\ref{tab:supp-lowpoly} extends the main-text surface-stage
comparison to additional point-cloud completion baselines, using the conversion and timing
protocol in Appendix~\ref{sec:supp_conversion}. All methods are evaluated
on the full 2,000-building BuildingWorld-Image test set, with shared
conversion settings within each backend.

The surface study in Table~\ref{tab:core-ablation}(c) is conducted in the point-cloud-input setting and uses the same predicted building structures as inputs to the reconstruction paths. The \emph{Structure only} variant converts these structures to low-poly meshes without the learned surface stage. The two \emph{+ Surface} variants insert surface generation before low-poly reconstruction. At inference, both surface generators are conditioned on the predicted occupied voxel coordinates, while the \emph{w/ point cond.} variant additionally receives point-cloud conditioning. This additional conditioning branch is included when training the corresponding surface generator. The \emph{w/o point cond.} variant uses only the predicted voxel coordinates at inference and is therefore not unconditional. PolyFit and RoLoPM are evaluated separately, following the fixed conversion protocols in Appendix~\ref{sec:supp_conversion}.

Adding surface generation reduces PolyFit CD from 10.10 to 9.06 and
increases F-score from 0.9387 to 0.9472. For RoLoPM, CD decreases from
9.83 to 8.65 and F-score increases from 0.9588 to 0.9719. The corresponding
time increases are 5.29\,s and 6.37\,s. The complete pipeline achieves
the lowest reported CD and highest F-score with each backend on this cohort.

Among methods using each conversion backend, the structure-only variant
has the shortest recorded time. The native BuildAnyPoint result takes
38.39\,s and is shown as a reference in both column groups. The reported
times reflect the combined costs of inference, adapter construction,
and low-poly conversion.

\paragraph{Additional point-cloud conditioning.}
Table~\ref{tab:core-ablation}(c) shows that additional point-cloud conditioning yields slightly worse reported geometry metrics with both backends. With PolyFit, $\mathrm{CD}_{L1}$ increases from 9.06 to 9.21 and F-score decreases from 0.9472 to 0.9458; with RoLoPM, the corresponding changes are 8.65 to 8.72 and 0.9719 to 0.9706. Recorded runtime also increases from 13.23 to 16.12 seconds with PolyFit and from 97.57 to 99.64 seconds with RoLoPM. These observations show no reconstruction benefit from injecting the point-cloud condition again at the surface stage in this experiment. The reported time combines inference and low-poly conversion and follows the timing convention in Appendix~\ref{sec:supp_conversion}; it does not isolate the execution time of the conditioning branch.

\paragraph{Qualitative surface comparison.}
In the surface-generation panel of the main-text component figure,
the Stage-1 support is held fixed.
The displayed Stage-2 outputs were selected from multiple
generation seeds to reduce visible surface defects;
the geometry itself was not filled or edited. These
are illustrative outputs, not an estimate of average
stochastic performance. The quantitative surface-stage results in
Table~\ref{tab:supp-lowpoly} use the recorded evaluation outputs,
not the seeds selected for display.

\begin{table*}[t]
\centering
\caption{\textbf{Low-poly reconstruction.} Arithmetic means on the
full 2,000-building BuildingWorld-Image test set, with the TopNet CD
exception specified below. CD denotes
$\mathrm{CD}_{L1}\times10^3$. F-scores are reported as fractions at a
normalized distance threshold of 0.02, to four decimal places for
Any2Building and two for the baselines. Time sums recorded inference
and conversion times.}
\label{tab:supp-lowpoly}
\small
\setlength{\tabcolsep}{6pt}
\begin{tabular}{lrrr rrr}
\toprule
& \multicolumn{3}{c}{PolyFit} & \multicolumn{3}{c}{RoLoPM} \\
\cmidrule(lr){2-4}\cmidrule(lr){5-7}
Method & Time (s)$\downarrow$ & CD$\downarrow$ & F-score$\uparrow$
& Time (s)$\downarrow$ & CD$\downarrow$ & F-score$\uparrow$ \\
\midrule
FoldingNet~\cite{foldingnet} & 121.29 & 48.02 & 0.42 & 239.08 & 51.76 & 0.31 \\
PCN~\cite{pcn} & 116.92 & 55.86 & 0.33 & 219.13 & 56.53 & 0.38 \\
TopNet~\cite{topnet}$^{\ddagger}$ & 217.59 & 87.90 & 0.09 & 382.77 & 64.25 & 0.21 \\
GRNet~\cite{grnet} & 125.69 & 70.88 & 0.33 & 215.34 & 54.77 & 0.30 \\
SnowflakeNet~\cite{snowflakenet} & 62.57 & 35.98 & 0.67 & 186.02 & 11.69 & 0.91 \\
PoinTr~\cite{pointr} & 73.33 & 42.55 & 0.58 & 204.71 & 12.68 & 0.88 \\
AnchorFormer~\cite{anchorformer} & 61.12 & 51.11 & 0.62 & 167.30 & 13.87 & 0.85 \\
AdaPoinTr~\cite{adapointr} & 60.91 & 40.14 & 0.62 & 161.57 & 11.05 & 0.92 \\
BuildAnyPoint~\cite{hua2026buildanypoint}$^{\dagger}$ & 38.39 & 12.88 & 0.92 & 38.39 & 12.88 & 0.92 \\
\midrule
Ours (Structure only) & \textbf{7.94} & 10.10 & 0.9387 & \textbf{91.20} & 9.83 & 0.9588 \\
Ours (Structure + Surface) & 13.23 & \textbf{9.06} & \textbf{0.9472} & 97.57 & \textbf{8.65} & \textbf{0.9719} \\
\bottomrule
\end{tabular}
\par\smallskip
\begin{minipage}{0.98\textwidth}
\footnotesize
Bold marks the best value in each column among methods using the
corresponding conversion backend. The surface generator in our full
pipeline uses only the predicted voxel coordinates as conditioning.
$^{\dagger}$Both column groups show the same native BuildAnyPoint
result as a reference.
$^{\ddagger}$PolyFit succeeds on approximately half of the TopNet outputs.
Failed conversions receive zero F-score. CD averages only successful
conversions, and the mean time includes all attempts.
Timing exclusions and adapters are specified in Appendix~\ref{sec:supp_conversion}.
\end{minipage}
\end{table*}

\section{Additional Ablations and Analysis}
\label{sec:supp_ablation}
This section examines point-cloud encoding, query resolution, and
geometric priors, and provides qualitative component comparisons and
textured visualizations. The encoder ablation uses Building-PCC;
the query-resolution experiment uses the full BuildingWorld-Image
test set. The paired point-degradation protocol is described in
Appendix~\ref{sec:supp_ablation_protocols}.

\subsection{Point-Cloud Encoder Ablation}
\label{sec:supp_encoder}
We compare different point-cloud conditioning encoders on
Building-PCC while keeping the spatial-query module and
generation pipeline fixed.
Specifically, we replace EdgeConv with a pointwise MLP,
a Point Transformer, or the TRELLIS.2 VAE
encoder~\cite{xiang2025trellis2}.
The point-based variants use the same sampled centers,
whereas the VAE encoder extracts conditioning features from
voxelized partial observations.
All encoder outputs are adapted to the same token count
and feature width.
We additionally remove the additive positional embedding
from the EdgeConv tokens to isolate its contribution
while retaining local neighborhood aggregation.
All variants follow the same training split, optimization
budget, and evaluation protocol.
Table~\ref{tab:encoder-ablation} compares point-cloud completion
and downstream low-poly reconstruction to assess the effects
of local geometry encoding, voxel-based conditioning,
and explicit point localization.

\begin{table}[t]
    \centering
    \caption{\textbf{Point-cloud encoder ablation on Building-PCC.}
    Metrics follow the Building-PCC comparison in the main text
    (PC: point cloud; LP: low-poly reconstruction).
    F-scores are reported as fractions rounded to two decimal places.
    Each variant replaces the point-cloud conditioning encoder
    or removes its additive positional embedding (PE).
    The spatial-query module, generative backbone, output decoder,
    and downstream reconstruction stages remain unchanged.}
    \label{tab:encoder-ablation}
    \footnotesize
    \setlength{\tabcolsep}{2.5pt}
    \resizebox{\columnwidth}{!}{%
    \begin{tabular}{lcccccc}
        \toprule
        Variant
        & PC F-score$\uparrow$
        & PC CD$\downarrow$
        & Uniformity$\downarrow$
        & EMD$\downarrow$
        & LP F-score$\uparrow$
        & LP CD$\downarrow$ \\
        \midrule
        TRELLIS.2 VAE encoder
        & 0.89 & 0.38 & 0.08 & 0.18 & 0.86 & 0.48 \\
        Pointwise MLP
        & 0.91 & 0.33 & 0.07 & 0.15 & 0.88 & 0.44 \\
        Point Transformer
        & 0.93 & 0.27 & 0.05 & 0.12 & 0.90 & 0.38 \\

        EdgeConv w/o PE
        & 0.95 & 0.22 & 0.04 & 0.09 & 0.93 & 0.33 \\
        \midrule
        EdgeConv (Ours)
        & 0.96 & 0.20 & 0.03 & 0.07 & 0.94 & 0.30 \\
        \bottomrule
    \end{tabular}%
    }
\end{table}

EdgeConv with positional embeddings has the highest PC and LP F-scores
and the lowest reported distance and Uniformity values among these
variants. Removing its positional embedding reduces PC F-score from
0.96 to 0.95 and LP F-score from 0.94 to 0.93. This comparison isolates
the additive center embedding while retaining neighborhood aggregation.

\subsection{Spatial-Query Design and Resolution}
\label{sec:supp_query_design}
To examine the spatial granularity of the query design,
we compare $8^3$ and $16^3$ query lattices through
controlled continuation from a common initialization,
matching the number of additional optimizer steps.
The coarser query field is interpolated to the generator's
$16^3$ latent grid, while the output resolution remains
$64^3$ in both variants. Query supervision and prior
aggregation follow the query resolution, so this
experiment evaluates the spatial-query configuration
as a whole rather than query count alone.

Both variants use EMA weights, 12 sampling steps,
guidance 3.0, and paired observations and noise on
the full 2,000-building BuildingWorld-Image test set.
Table~\ref{tab:supp-query-resolution} reports voxel F1
for point-only (P), image-only (I), joint point--image
(P+I), and image-plus-prior (I+R) conditioning.

\begin{table}[t]
\centering
\caption{Query-resolution comparison on the full 2,000-building
BuildingWorld-Image test set. Voxel F1 measures surface-support overlap;
higher is better.}
\label{tab:supp-query-resolution}
\small
\setlength{\tabcolsep}{6pt}
\renewcommand{\arraystretch}{1.12}
\begin{tabular}{@{}lcccc@{}}
\toprule
Query grid & P & I & P+I & I+R \\
\midrule
$8^3$  & 0.7703 & 0.2213 & 0.7671 & 0.2681 \\
$16^3$ & 0.7723 & 0.2272 & 0.7676 & 0.2987 \\
\bottomrule
\end{tabular}
\end{table}

The finer lattice yields slightly higher mean voxel F1
in all four modes, with the largest observed gain under
image-plus-prior conditioning. For $16^3-8^3$, the mean
paired differences are +0.0020 (P), +0.0059 (I),
+0.0004 (P+I), and +0.0306 (I+R). Differences are computed
before rounding and may therefore differ slightly from
subtraction of the displayed table entries.

\subsection{Image-Conditioned Geometric Priors}
\label{sec:supp_geometric_prior}
\label{sec:supp_prior}
\paragraph{Prior-assisted generation on Skylines-50K.}
Following the 500-asset evaluation setting of
SatSkylines~\cite{jin2026satskylines}, we reproduce the
compared methods on Skylines-50K and evaluate LoD0 and
LoD1 priors separately.
As shown in Table~\ref{tab:prior-generation}, Any2Building
achieves the best results across all geometry and appearance
metrics under both prior settings, demonstrating effective
building reconstruction with geometric constraints of
varying detail.
\begin{table}[t]
    \centering
    \caption{\textbf{Prior-assisted building generation with
    LoD0 and LoD1 priors.}
    Results follow the 500-asset evaluation setting and
    are compared separately within each prior level.
    F-scores are reported as fractions. PSNR is reported in dB.}
    \label{tab:prior-generation}
    \footnotesize
    \setlength{\tabcolsep}{3pt}
    \renewcommand{\arraystretch}{1.1}
    \resizebox{\columnwidth}{!}{%
    \begin{tabular}{@{}lrrrrrr@{}}
        \toprule
        Method
        & \multicolumn{3}{c}{Geometry}
        & \multicolumn{3}{c}{Appearance} \\
        \cmidrule(lr){2-4}
        \cmidrule(lr){5-7}
        & IoU$\uparrow$
        & CD$\downarrow$
        & F-score$\uparrow$
        & PSNR$\uparrow$
        & LPIPS$\downarrow$
        & CLIP$\uparrow$ \\
        \midrule
        \multicolumn{7}{l}{\textit{(a) LoD0 priors}} \\
        SpaceControl~\cite{fedele2026spacecontrol}
        & 0.6318 & 0.0402 & 0.5924
        & 16.2318 & 0.2714 & 0.6253 \\
        Hunyuan3D-Omni~\cite{hunyuan2025omni}
        & 0.8705 & 0.0268 & 0.7920
        & 17.4162 & 0.2187 & 0.7468 \\
        SatSkylines~\cite{jin2026satskylines}
        & 0.8982 & 0.0224 & 0.8236
        & 17.9248 & 0.2015 & 0.7601 \\
        Ours
        & \textbf{0.9307} & \textbf{0.0172} & \textbf{0.8702}
        & \textbf{19.5238} & \textbf{0.1624} & \textbf{0.7889} \\
        \midrule
        \multicolumn{7}{l}{\textit{(b) LoD1 priors}} \\
        SpaceControl~\cite{fedele2026spacecontrol}
        & 0.6612 & 0.0357 & 0.6218
        & 16.8712 & 0.2481 & 0.6432 \\
        Hunyuan3D-Omni~\cite{hunyuan2025omni}
        & 0.8894 & 0.0231 & 0.8107
        & 18.0512 & 0.1986 & 0.7590 \\
        SatSkylines~\cite{jin2026satskylines}
        & 0.9165 & 0.0189 & 0.8421
        & 18.6374 & 0.1813 & 0.7742 \\
        Ours
        & \textbf{0.9452} & \textbf{0.0139} & \textbf{0.8906}
        & \textbf{20.4162} & \textbf{0.1402} & \textbf{0.8015} \\
        \bottomrule
    \end{tabular}%
    }
\end{table}

\begin{figure}[t]
\centering
\includegraphics[width=\columnwidth]{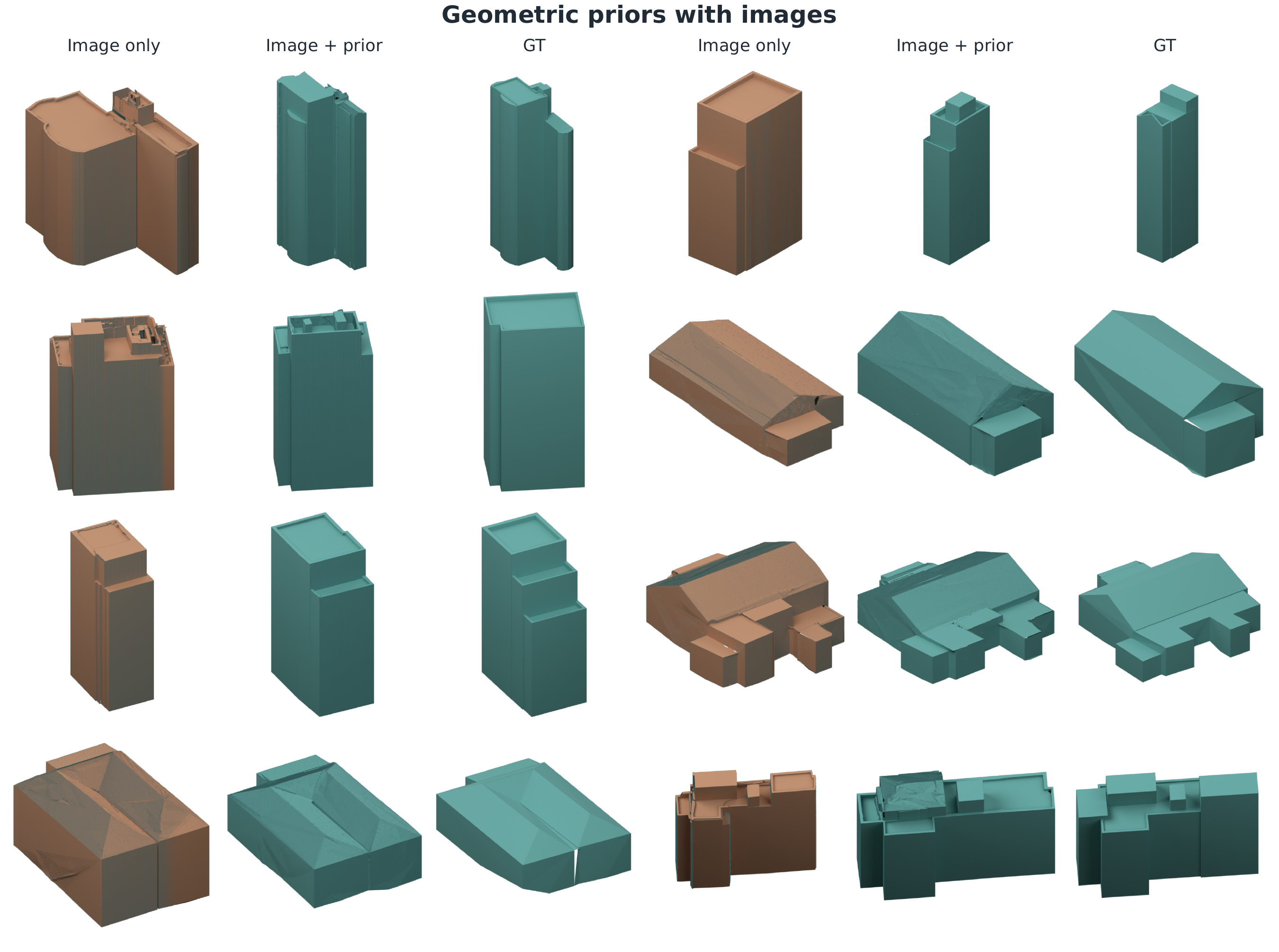}
\caption{\textbf{Geometric priors reduce image-to-3D ambiguity.}
Selected image-only and image-plus-prior reconstructions,
each shown with its reference geometry. Priors help
anchor building extent and massing while the generator
recovers architectural detail. These examples illustrate
the additional-input setting rather than image-only
performance.}
\label{fig:supp-prior}
\end{figure}

A rough prior constrains the feasible building volume
while leaving the generator to infer roof structure
and facade articulation. Figure~\ref{fig:supp-prior}
compares image-only generation with prior-assisted
generation on selected examples. The examples show roof structure beyond the coarse prior, but do not
quantify a mean improvement over a matched image-only baseline.

Prior quality and provenance matter: target-derived
heights constitute extra supervision at inference,
and incorrect footprints or heights can bias the result.
We therefore distinguish the prior-conditioned setting
from image-only reconstruction throughout.

\subsection{Qualitative Interpretation of Components}
\label{sec:supp_qualitative}
\begin{figure*}[p]
\centering
\includegraphics[
  width=\textwidth,
  height=.87\textheight,
  keepaspectratio
]{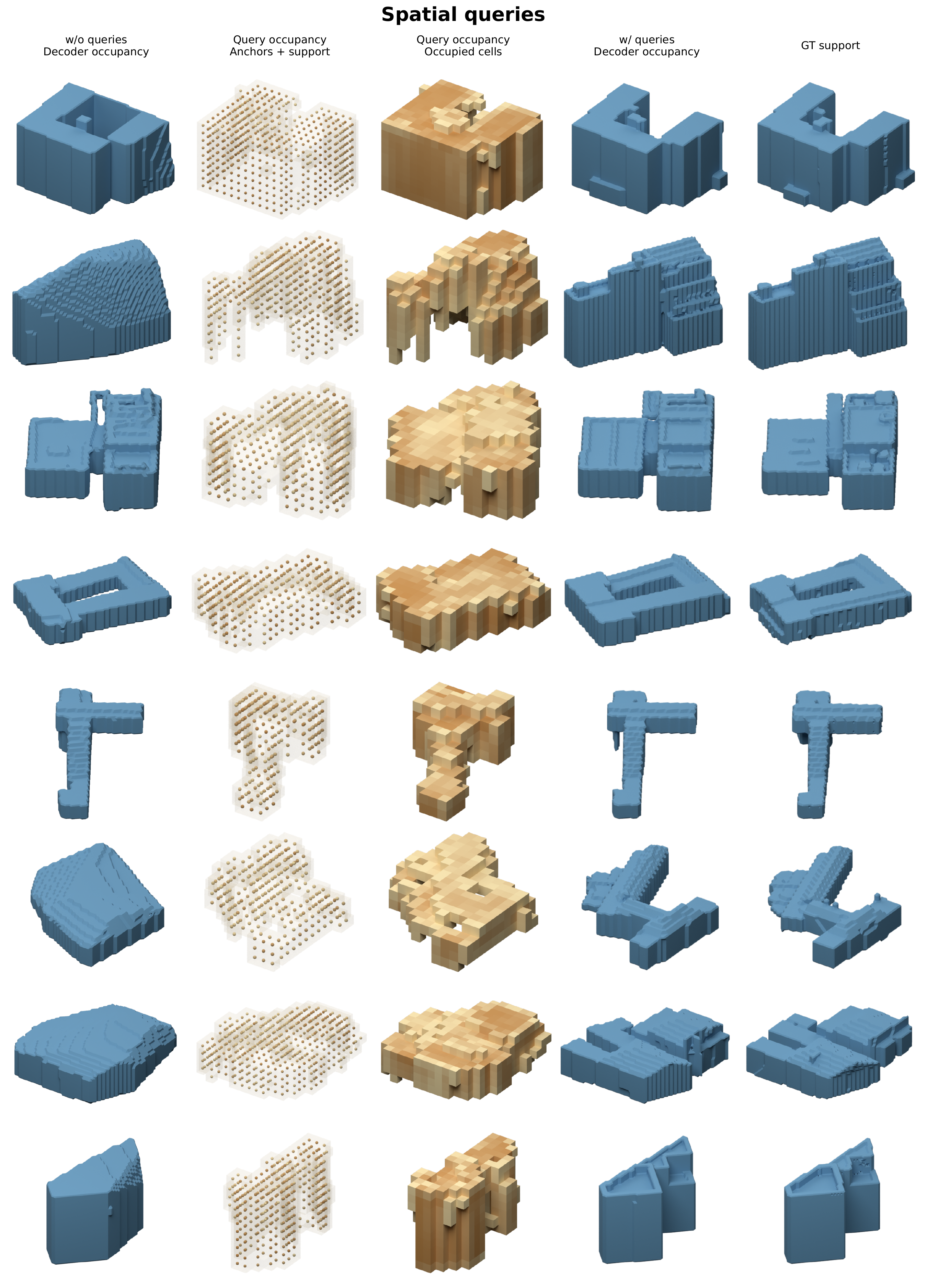}
\caption{\textbf{Spatial queries provide an interpretable
geometric intermediate.}
From left to right: decoder support without queries,
query anchors with occupancy support, occupied query
cells, decoder support with queries, and reference
support. The query lattice is $16^3$, with occupied
cells visualized at probability at least 0.5.
Each row preserves a common view. Quantitative component comparisons
are reported in Table~\ref{tab:core-ablation}(a).}
\label{fig:supp-query}
\end{figure*}

Figure~\ref{fig:ablation}(b) and Fig.~\ref{fig:supp-query}
visualize query occupancy and decoded building support alongside
the reference surface voxels. Figure~\ref{fig:ablation}(c) shows
predictions from single-task and jointly trained models.
These examples complement the quantitative component comparisons
in Table~\ref{tab:core-ablation}(a).
The surface-generation comparison in Figure~\ref{fig:ablation}(a)
is discussed in Appendix~\ref{sec:supp_lowpoly}.

\subsection{Downstream Textured Visualization}
\label{sec:supp_texture}
\begin{figure*}[t]
\centering
\includegraphics[width=\textwidth]{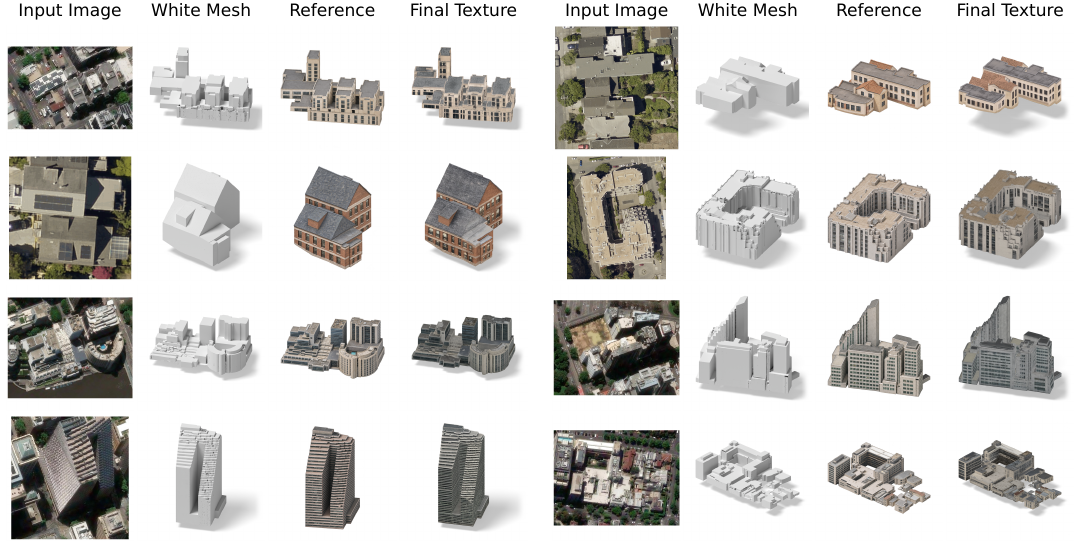}
\caption{\textbf{Downstream textured building visualization.}
Each group shows the input image, untextured mesh (White Mesh),
reference appearance, and final textured output. The examples
illustrate the appearance of the textured buildings alongside
their geometric structure.}
\label{fig:supp-texture}
\end{figure*}

Figure~\ref{fig:supp-texture} illustrates downstream textured
visualization, pairing input imagery and untextured building
meshes with reference appearances and final textured outputs.
These qualitative examples complement the reconstruction results in
Appendix~\ref{sec:supp_lowpoly}.

\end{document}